\documentclass[journal,table]{nesy2025} 
\usepackage[utf8]{inputenc}
\usepackage{float}
\usepackage{rotating}
\usepackage{placeins}
\usepackage{rotating}
\usepackage{longtable}
\usepackage{booktabs}
\usepackage[load-configurations=version-1]{siunitx} 

\usepackage{tikz}
\usepackage{xcolor}

\definecolor{strucmol}{HTML}{54278F}
\definecolor{smallmolsens}{HTML}{006D2C}
\definecolor{molfuncreg}{HTML}{6F6F6F}

\makeatletter
\DeclareRobustCommand{\pdot}{\mathbin{\mathpalette\pdot@\relax}}
\newcommand{\pdot@}[2]{%
  \ooalign{%
    $\m@th#1\circ$\cr
    \hidewidth$\m@th#1\cdot$\hidewidth\cr
  }%
}
\makeatother

\theorembodyfont{\upshape}
\theoremheaderfont{\scshape}
\theorempostheader{:}
\theoremsep{\newline}

\def\thickhline{%
  \noalign{\ifnum0=`}\fi\hrule \@height \thickarrayrulewidth \futurelet
  \reserved@a\@xthickhline}
  
\def\@xthickhline{\ifx\reserved@a\thickhline
  \vskip\doublerulesep
  \vskip-\thickarrayrulewidth
    \fi
  \ifnum0=`{\fi}}

\newlength{\thickarrayrulewidth}
\title[]{Graph Neural Network based Hierarchy-Aware Embeddings of Knowledge Graphs: Applications to Yeast Phenotype Prediction\texorpdfstring{\protect\footnotemark[2]}{}}

 \author{\Name{Filip Kronström} \Email{filipkro@chalmers.se}\\
 \Name{Alexander H. Gower}\\
 \Name{Daniel Brunnsåker}\\
 \addr Department of Computer Science and Engineering,\\Chalmers University of Technology and University of Gothenburg, Sweden
 \AND
 \Name{Ievgeniia A. Tiukova}\\
 \addr Department of Life Sciences, Chalmers University of Technology, Sweden\\
 \addr Department of Industrial Biotechnology, KTH Royal Institute of Technology, Sweden
 \AND
 \Name{Ross D. King}\\
 \addr Department of Computer Science and Engineering,\\Chalmers University of Technology and University of Gothenburg, Sweden\\
 \addr Department of Chemical Engineering and Biotechnology, University of Cambridge, United Kingdom
 }

 \newcommand{\norm}[1]{\left\lVert#1\right\rVert}

\usepackage{hyperref}
\begin{document}
\newcommand{\colbox}[2]{\colorbox{#1}{\parbox{0.9\textwidth}{#2}}}
\maketitle
\begingroup
  \renewcommand\thefootnote{$\dagger$} 
  \footnotetext[2]{This is an extended version of the paper ``Ontology-based box embeddings and knowledge graphs for predicting phenotypic traits in \emph{Saccharomyces~cerevisiae}''~\citep{kronstrom2025ontologybased}, presented at NeSy 2025.}
\endgroup

\begin{abstract}

We present a method for finding hierarchy-aware embeddings of knowledge graphs (KGs) using graph neural networks (GNNs) enriched with a semantic loss derived from underlying ontologies. This method yields embeddings that better reflect domain knowledge. To demonstrate their utility, we predict and interpret the effects of gene deletions in the yeast \emph{Saccharomyces~cerevisiae} and learn box embeddings for KGs in the absence of a prediction task. We further show how box embeddings can serve as the basis for evaluating KG revisions.

Our yeast KG is constructed from community databases and ontology terms.
Low-dimensional box embeddings combined with GNNs
are used to predict cell growth for double gene knockouts. Over 10-fold cross validation, these predictions have a mean $R^2$~score~of~0.360, significantly higher than baseline comparisons, demonstrating that high-level qualitative knowledge is informative about experimental outcomes.
Incorporating semantic loss terms in the training of the models improves their predictive performance \nobreak{($R^2$=0.377)} by aligning embeddings with ontology structure. This shows that class hierarchies from ontologies can be exploited for quantitative prediction.
We also test the trained models on triple gene knockouts, showing they generalise to data beyond those seen in training.


Additionally, by identifying co-occurring relations in the yeast KG important for the cell-growth predictions, we construct hypotheses about interacting traits in yeast. A biological experiment validates one such finding, revealing an association between inositol utilisation and osmotic stress resistance, highlighting the model's potential to guide biological discovery.

\end{abstract}

\section{Introduction}
Knowledge graphs (KGs) are widely used to represent structured knowledge as sets of triples of the form (\textit{subject, predicate, object}). In many domains, including the life sciences, KGs are enriched with ontological information, where formally defined vocabularies describe classes of entities and relations between them. In particular, hierarchical\footnote{Section~\ref{sec:box_gnn} provides a detailed definition of what is meant by the terms \textit{hierarchy} and \textit{hierarchical} in this work.} class information expressed through the `\texttt{subClassOf}' relation is prevalent, as domain experts can organise concepts, without requiring formal expertise in logic or ontology engineering. As a result, such hierarchies constitute a pragmatic and high-impact source of background knowledge for representation learning.

Representation learning enables the transformation of KGs into a form that works as an input to a program class that does not accept the KG itself as an input. A form of representation learning that has proven useful for downstream tasks, such as link prediction and property prediction, is embedding KGs into an $n$-dimensional vector space. Many KG embedding (KGE) methods are trained primarily on observed triples, using the relational structure of the graph as the main learning signal. However, when ontological knowledge is present, an embedding should ideally reflect not only the connectivity of the graph but also background semantic constraints. Rewarding compatibility between learned representations and known domain structure can provide additional inductive bias during training, improve generalisation beyond observed data, and support more interpretable predictions~\citep{gutierrez-basulto_knowledge_2018}.



Incorporating semantic constraints into continuous representations raises the question of how ontological concepts should be represented geometrically. Hierarchical relations impose inclusion constraints, which can be captured using point embeddings in non-Euclidean spaces, such as hyperbolic embeddings~\citep{nickel_poincare_2017}. However, when the goal is to explicitly encode class inclusion and concept-level constraints, volumetric representations provide a natural alternative. In this paradigm, concepts are modelled as subsets of a latent space using geometric objects, such as hyperspheres~\citep{kulmanov_embeddings_2019}, hypercones~\citep{ganea_hyperbolic_2018}, or hyperrectangles (boxes)~\citep{vilnis_probabilistic_2018,peng_description_2022,xiong_faithful_2022,jackermeier_dual_2024,yang_transbox_2025}, enabling hierarchical relations to be expressed directly through geometric containment.

While volumetric representations are well suited for capturing ontological structure, it is challenging to integrate them with complex heterogeneous graphs, and to train them end-to-end for task-specific prediction. Many real-world KGs are large, heterogeneous, and feature-rich. This motivates the use of graph neural networks (GNNs) as a framework for representation learning through aggregation of information from local neighbourhoods and node attributes~\citep{kipf_semi-supervised_2017, hamilton_inductive_2017}. GNN-based approaches have been successfully applied to KGs and relational data for tasks such as node classification, link prediction, and property prediction~\citep{schlichtkrull_modeling_2018, ye_comprehensive_2022}.

However, standard GNN architectures do not explicitly enforce global semantic constraints derived from ontologies, and may therefore learn representations that violate known hierarchical relationships, even when such information is available as background knowledge. This motivates approaches that combine data-driven representation learning with explicit semantic constraints derived from ontologies. We also hypothesise that, particularly in settings with low-dimensional or scarce data, that the inductive bias introduced by using ontology constraints during representation learning will improve performance on downstream prediction tasks.

In this paper, we address this limitation by combining GNN-based KG embeddings with box-based representations of ontological concepts by introducing Hierarchy-aware GNNs. Our methodological contribution is a framework in which box embeddings are used to encode hierarchical background knowledge, while GNNs are used to learn features from the graph structure. 



We evaluate this approach in the context of biological knowledge about the yeast \emph{Saccharomyces cerevisiae}. Yeast is among the most extensively studied model organisms and plays a central role in both basic research and industrial applications~\citep{parapouli_saccharomyces_2020}. Decades of experimental work have produced large amounts of structured data, available through curated databases such as the Saccharomyces Genome Database~\citep{engel_saccharomyces_2024}. These resources integrate experimental observations with ontological annotations describing biological processes, phenotypes, chemical compounds, and interactions.

Despite the depth of scientific research on yeast, our understanding of its biology remains incomplete: many genes are poorly annotated, and current models fail to predict many phenotypic outcomes of complex genetic interactions~\citep{wood_hidden_2019,costanzo_global_2019}. Experimental investigation is therefore essential for advancing biological knowledge. However, biological experimentation is costly, and the space of possible experiments is vast. As a result, computational methods that support hypothesis generation at scale are of high value~\citep{king_functional_2004,brunnsaker_agentic_2025}.


We construct an ontology-enriched KG describing yeast biology and apply our embedding framework to predict gene fitness and to generate hypotheses about interactive properties. This application serves both as a realistic use case and as an empirical evaluation of whether incorporating hierarchical semantic constraints into KG embeddings improves predictive performance and supports biologically meaningful knowledge discovery.


The remainder of the paper is structured as follows.
Section~\ref{sec:rel_work} presents related work relevant for this paper. Sections~\ref{sec:background}~and~\ref{sec:preliminaries} introduce background, and define terms and losses used throughout the paper. In Section~\ref{methods} we present Hierarchy-aware GNNs, a \emph{Saccharomyces cerevisiae} KG, and applications of the Hierarchy-aware GNNs on this KG. These results are presented in Section~\ref{sec:results}, and discussion and future work can be found in Section~\ref{sec:discussion}.





\section{Related work}
\label{sec:rel_work}


Knowledge graph embedding methods aim to represent entities and relations in a continuous vector space such that observed facts are preserved geometrically. In their most simple formulation, KGs are treated as collections of triples of the form (\textit{subject, predicate, object}) or (\textit{head, relation, tail}), and embeddings are learned by optimising a scoring function over such triples.

One approach is to model relations using bilinear interactions between entity embeddings. RESCAL~\citep{nickel_three-way_2011} represents each relation as a full matrix, enabling expressive modelling of many-to-many relations but incurring high computational cost. DistMult~\citep{yang_embedding_2015} simplifies this formulation by restricting relation matrices to be diagonal, yielding a model that performs well for symmetric relations but cannot capture asymmetry. ComplEx~\citep{trouillon_complex_2016} extends DistMult to a complex-valued space, allowing asymmetric relations to be modelled while retaining computational efficiency.

Another prominent family of approaches models relations as transformations applied to entity embeddings. TransE~\citep{bordes_translating_2013} represents relations as translations between entity vectors such that $s + p \approx o$, offering simplicity and interpretability, but struggling with complex relations. RotatE~\citep{sun_rotate_2019} generalises this idea by interpreting relations as rotations in the complex plane, enabling the modelling of symmetry, antisymmetry, inversion, and composition while achieving strong empirical performance.

\paragraph*{Incorporating hierarchical information.}

Despite differences in formulation and expressivity, the models mentioned so far share a common focus on instance-level relational structure derived from observed triples. While they can capture relational patterns and implicit structural regularities, ontological constraints such as class subsumption or disjointness are typically not enforced explicitly. This has motivated extensions of KGE methods that seek to incorporate hierarchical information while preserving the efficiency and scalability of triple-based learning.



One such approach is HAKE~\citep{zhang_learning_2020}, which augments standard KG embeddings with hierarchy awareness. It introduces a polar coordinate representation in which hierarchical depth and lateral similarity are modelled separately, enabling improved link prediction in graphs with pronounced hierarchical structure.
Conceptually, HAKE shares similarities with hyperbolic and Poincaré embedding approaches in how it captures hierarchical structure, despite operating in a Euclidean space. While these approaches capture hierarchical patterns through geometry, they do not explicitly represent ontological concepts or logical axioms.

A semantics-driven line of work focuses on embedding ontologies formulated in description logics by constructing geometric representations that approximate model-theoretic interpretations. ELEm~\citep{kulmanov_embeddings_2019} introduces a volumetric embedding approach in which classes and nominals are represented as hyperspheres, and relations are modelled using TransE-style translations. Subsumption and existential restrictions are enforced via geometric containment, demonstrating that $\mathcal{EL}^{++}$ axioms can be satisfied approximately in a continuous space.

ELBE~\citep{peng_description_2022} builds directly on earlier volumetric ontology embeddings by replacing spherical concept regions with axis-aligned hyperrectangles (boxes), ensuring closure under intersection. This geometric choice enables a more faithful encoding of $\mathcal{EL}^{++}$ axioms while retaining a volumetric interpretation of concepts. Subsequent work further extends box-based ontology embeddings within $\mathcal{EL}^{++}$: BoxEL~\citep{xiong_faithful_2022} represents concepts as boxes and roles as affine transformations, providing soundness guarantees with respect to $\mathcal{EL}^{++}$-semantics; Box$^2$EL~\citep{jackermeier_dual_2024} represents both concepts and roles as boxes to support many-to-many relations; and TransBox~\citep{yang_transbox_2025} provides an $\mathcal{EL}^{++}$-closed construction that enables compositional modelling of complex concepts and many-to-many roles through box-based sets of translations.


To address issues of noise and uncertainty, probabalistic approaches to KG embeddings have also been developed, where embedding parameters are modelled as random variables. \cite{vilnis_probabilistic_2018} introduces probabilistic box representations to model uncertainty and inclusion relationships, while~\cite{dasgupta_improving_2020} addresses optimisation and identifiability issues arising in such models. 


\paragraph*{Graph neural network approaches.}

While the approaches above focus on how entities, relations, or concepts are represented geometrically, they typically learn embeddings either in isolation or directly from triples or axioms. In contrast, many real-world knowledge graphs are large and heterogeneous, motivating methods that explicitly aggregate information from local graph structure. Graph neural networks (GNNs) provide such a framework by learning node representations through iterative message passing over graph neighbourhoods, naturally supporting end-to-end optimisation for specific downstream prediction tasks.

Relational Graph Convolutional Networks (R-GCNs)~\citep{schlichtkrull_modeling_2018} were developed specifically with knowledge graphs in mind and have been applied to tasks such as link prediction and node classification. R-GCNs extend standard GCNs to multi-relational settings by introducing relation-specific message transformations, an idea that can be combined with a wide range of message-passing architectures.

More generally, the integration of symbolic knowledge into neural models has been explored through semantic loss functions, which penalise predictions that violate logical constraints during training~\citep{xu_semantic_2018}. While originally proposed for feed-forward neural networks, this idea provides a general mechanism for incorporating prior knowledge into end-to-end learning.

Box embeddings have also been combined with GNNs in specific application settings. BoxGNN~\citep{lin_when_2024} integrates box-based representations into a GNN architecture for recommendation tasks by redefining message passing and aggregation operations directly in box space, using geometric operations such as intersection and union. In this setting, boxes serve as the primary representation for users, items, and attributes. This approach is tailored to recommendation and does not aim to enforce semantic constraints in knowledge graph embeddings.




\paragraph*{Knowledge graph embeddings in the natural sciences.}

KGs have been widely adopted in the biomedical domain as a means of integrating heterogeneous experimental and curated data into a unified representation.
For example BioKG~\citep{walsh_biokg_2020} and SPOKE~\citep{morris_scalable_2023} combine information from different databases to create one large heterogeneous graph with information about, for example, genes and drugs. There are also graphs describing more narrow phenomena such as the protein-protein associations and the drug-drug interactions in the Open Graph Benchmark~\citep{hu_open_2020}.

\cite{memariani_box_2025} propose a box-embedding framework for extending the ChEBI chemical ontology from molecular data. In their model, chemical classes are represented as boxes, and molecules, encoded from SMILES strings using an ELECTRA-based encoder~\citep{clark_electra_2019}, are represented as points. Molecules are trained to lie inside the boxes of their annotated classes (including all superclasses), so that box containment and overlap reflect subsumption and disjointness relations between classes. This enables evaluation not only of molecular classification, but also of how well the learned box geometry recovers the class taxonomy.

\sloppy
Predicting biological properties from structured background knowledge can be approached in several ways. \cite{ma_using_2018} encode GO-annotations together with the GO hierarchy in a neural network to predict cellular growth in \emph{S.~cerevisiae}. By mining patterns from a Datalog knowledge base containing facts from databases, \cite{brunnsaker_interpreting_2024} connect qualitative biological concepts to quantitative intracellular protein abundance measurements. \cite{gualdi_predicting_2024} predict gene-disease associations using embeddings derived from a protein interaction knowledge graph, evaluating a range of methods including translational KG embeddings (e.g. TransE and RotatE), random-walk-based approaches, and GNN-based models. In their approach, embeddings are learned independently of the prediction task and subsequently used as features for supervised classifiers such as support vector machines and tree-based models, resulting in a two-stage pipeline rather than end-to-end training.

\section{Background}
\label{sec:background}


\subsection*{Graph neural networks}

Graph neural networks (GNNs) are neural network models that take graphs as inputs, and learn vector representations of nodes. In a GNN, the structure of the graph informs the architecture, with information being passed from the neighbourhood of each node. At each message passing layer of the GNN, information is aggregated from a node's neighbourhood to update its representation.

Each message passing layer therefore propagates information from the immediate neighbourhood of a node, so by increasing the number of message passing layers we increase the distance in the graph that information propagates---this is referred to as the receptive field of the GNN. Message passing layers are often interspersed with pooling layers to increase the receptive field. An example of a message aggregation scheme is a convolutional layer, the basic component of a graph convolutional network (GCN). In a GCN layer, updates to node embeddings at each layer are calculated by multiplying the embeddings from the previous layer by a scaled adjacency matrix (including self-adjacency) and a weight matrix~\citep{kipf_semi-supervised_2017}.

While a very powerful tool, particularly when learning representations for homogeneous graphs, GCNs do not use information about edge type when learning on heterogeneous graphs, which most KGs are. R-GCNs apply the same ideas as GCNs, but instead allow for separate weight matrices depending on edge type, and therefore are much better suited to representation learning on heterogeneous KGs than GCNs~\citep{schlichtkrull_modeling_2018}.


\subsubsection*{GraphSAGE}

GraphSAGE~\citep{hamilton_inductive_2017} is an alternative GNN framework for learning node embeddings that, instead of directly learning embeddings for nodes in a static graph, learns a function that can generate embeddings. As a result, GraphSAGE is able to generate embeddings for new nodes that were not in the graph the model was trained on, provided these new nodes share the same attribute schema as the original graph. GraphSAGE uses an aggregator to collect information from the neighbourhood of a given node; concatenates this aggregated information with the current embedding state of the node; and then passes this through a weighted function to propagate the information. The aggregator can be directly encoded, for example a simple mean, or it could be learned during training. The receptive field of the network can be increased by adding more layers.


\subsection*{Description logics}

Description logics (DLs) are fragments of first-order logic that allow only for constants (individuals), unary predicates (concepts), and binary predicates (roles). The symbol $\sqsubseteq$ is used to represent concept inclusion, and $\equiv$ is used for equivalence. We use DL to express axioms to describe a domain, resulting in a knowledge graph. These axioms are generally split into terminological axioms (TBox)---regarding general knowledge about concepts and roles in the domain---and assertional axioms (ABox)---which make statements about individuals~\citep{baader_introduction_2017}. For example, we can represent TBox axioms for the statements: (\ref{eq:tbox-prot}) ``proteins are molecules''; (\ref{eq:tbox-carb}) ``carbohydrates are molecules''; (\ref{eq:tbox-disj}) ``carbohydrates are disjoint from proteins''; and (\ref{eq:tbox-enz}) ``enzymes are molecules that catalyze a biochemical reaction'' in DL:
\begin{align}
  \texttt{Protein} &\sqsubseteq \texttt{Molecule}\label{eq:tbox-prot} \\
  \texttt{Carbohydrate} &\sqsubseteq \texttt{Molecule}\label{eq:tbox-carb} \\
  \texttt{Carbohydrate} \sqcap \texttt{Protein} &\sqsubseteq \bot\label{eq:tbox-disj} \\
  \texttt{Enzymes} &\equiv \texttt{Molecule} \sqcap \exists\texttt{Catalyzes}{.}\texttt{Reaction}\label{eq:tbox-enz}
\end{align}
\noindent Two examples of ABox axioms are:
\begin{align*}
  &\texttt{Protein(hexokinase)}. \\
  &\texttt{Catalyzes(hexokinase,glucose\_phosphorylation)}.
\end{align*}
\noindent From the TBox and ABox, we could deduce that hexokinase is an enzyme, because it catalyzes the phosphorylation of glucose.

Individuals, concepts, and roles in DLs can be represented in Web Ontology Language (OWL) using individual, class, and property statements.

$\mathcal{EL}$++ is a lightweight DL designed for efficient reasoning over large ontologies, allowing conjunctions, existential restrictions, and role hierarchies while ensuring polynomial-time reasoning~\citep{baader_introduction_2017}. It underpins the OWL 2 EL profile and is widely used in biomedical ontologies, including the ones introduced below.


\subsection*{Ontologies and data}


Decades of research on \emph{Saccharomyces cerevisiae} have resulted in extensive knowledge that is available both in the scientific literature and in curated biological databases. The Saccharomyces Genome Database (SGD)~\citep{engel_saccharomyces_2024} provides a central resource aggregating curated information about \emph{S. cerevisiae} genes, including Gene Ontology annotations, experimentally observed phenotypes, and regulatory as well as genetic interaction data. Complementary information about biochemical reactions, events in which substrates are transformed into products, and pathways, sets of interconnected reactions driving cellular functions, is available in pathway-oriented resources such as BioCyc~\citep{karp_biocyc_2019}.

The contents of these databases are represented using ontologies and controlled vocabularies to ensure consistency and interoperability. The Gene Ontology (GO)~\citep{ashburner_gene_2000} defines classes describing molecular functions, biological processes, and cellular components. Phenotypic information in SGD is formalised using the Ascomycete Phenotype Ontology (APO)~\citep{costanzo_new_2009}, which captures observable characteristics arising from interactions between genotype and environment, such as growth defects or resistance to chemical perturbations. Chemical compounds are described using Chemical Entities of Biological Interest (ChEBI)~\citep{hastings_chebi_2016}. Genetic, physical, and regulatory interactions are modelled using the Interaction Network Ontology (INO)~\citep{hur_development_2015} and Molecular Interactions (MI)~\citep{hermjakob_hupo_2004}. Commonly used relations shared across ontologies are defined in the Relations Ontology (RO)~\citep{mungall_oborelobo-relations_2020}. Among these resources, only GO and ChEBI define relations between classes, whereas the remaining ontologies provide taxonomies of domain-specific entities.

\section{Preliminaries}
\label{sec:preliminaries} 
\subsection*{Box embeddings}

Axis-aligned hyperrectangles, or ``boxes'' as they often are referred to, are defined as the Cartesian product of closed intervals,
\begin{equation}
  \text{Box} = \prod_{i=1}^n[z_i, Z_i],
\end{equation}
where $z_i$ and $Z_i$ correspond to the lower and upper coordinate along dimension $i$. To fulfil the criteria that the upper coordinate should be greater than or equal to the lower coordinate, $Z_i \geq z_i$, we create boxes from latent variables, $\theta$, using the \texttt{MinDeltaBoxTensors} constructor introduced by \cite{chheda_box_2021}. The upper and lower box coordinates are defined as follows:
\begin{equation}
  z_i(\theta_i) = \theta_i^z, \qquad Z_i(\theta_i) = z_i + \text{softplus}(\theta_i^Z).
  \label{eq:box_transform}
\end{equation}
\noindent The equation for transforming latent variables, $\theta$ into box representations is therefore:
\begin{equation}
  \text{Box}(\theta) = \prod_{i=1}^n[z_i(\theta_i), Z_i(\theta_i)]
  \label{eq:box_transform_full}
\end{equation}
Boxes can also be represented by their centre-point, $c_i$, and offset, $o_i$, along dimension $i$, found from $z$ and $Z$ as follows:
\begin{equation}
  c_i = \frac{z_i + Z_i}{2}, \qquad o_i = Z_i - c_i
\end{equation}

A knowledge graph (KG) can be represented using box embeddings by embedding classes and individuals (nodes) as axis-aligned boxes in a low-dimensional space. As discussed in Section~\ref{sec:rel_work}, this idea has been explored in several forms, with its popularity largely driven by a favourable trade-off between expressivity and computational efficiency. Central to box embeddings is the modelling of transitive relations through geometric containment, where the box of a head entity is constrained to lie within the box of a tail entity. In this work, we use box embeddings to represent `\texttt{subClassOf}' relationships between classes, rewarding containment of each subclass box within its corresponding superclass box.

\subsection*{Semantic losses}
To learn box embeddings we consider two different types of loss functions for concept inclusions of the form $C \sqsubseteq D$. The first loss, here called $\mathcal{L}_{distance}$, has previously been used by, for example, \cite{peng_description_2022} and \cite{jackermeier_dual_2024}. Using the nomenclature presented above it is calculated by first finding the element-wise distance between the two boxes,

\begin{equation}
  d(C_i,D_i) = |c_i^C - c_i^D| - o_i^C - o_i^D
  \label{eq:box_distance}
\end{equation}
In the loss function, this quantifies how far a subclass box is from being completely contained within the superclass box,
\begin{equation}
  \mathcal{L}_{distance}(C, D) = \Big|\Big|\Big(\max(0,d(C_i, D_i) + 2o_i^C)\Big)_{i=1}^n\Big|\Big|
  \label{eq:dist_loss}
\end{equation}
Note here that \cite{jackermeier_dual_2024} used the loss in \eqref{eq:dist_loss} in the case $D\neq\emptyset$, and a separate loss if $D=\emptyset$. In this work, we don't consider that it is reasonable to assume that any of the concepts included in the knowledge graph is empty. Mostly, this is because the concepts in the knowledge graphs used here come directly from scientific ontologies, where their inclusion implicitly represents a belief of their existence, but also relies on an open world assumption. It could be that the loss as defined in \cite{jackermeier_dual_2024} is necessary for complex or compound concepts in other applications, where the empty set is a possibility for concept $D$. We also omit the margin parameter from \cite{jackermeier_dual_2024} as we want the loss to be zero when the subclass axiom is satisfied in the embedding space.
To keep disjoint classes, $C \sqcap D \sqsubseteq \bot$, apart we penalise overlap of the boxes by the following loss (where $\mathbb{I}$ is the indicator function):

\begin{equation}
  \mathcal{L}_{distance}^-(C, D) = \Big|\Big|\Big(\max(0,-d(C_i, D_i))\Big)_{i=1}^n\Big|\Big|\cdot\prod_{i=1}^n\mathbb{I}\big[d(C_i, D_i)<0\big]
  \label{eq:dist_loss_neg}
\end{equation}

The second loss type considers the overlap between boxes and for the subsumption $C \sqsubseteq D$ it is calculated as
\begin{equation}
  \mathcal{L}_{overlap} = -\log\Big(\frac{\textsf{Vol}(\text{Box}(C) \cap \text{Box}({D}))}{\textsf{Vol}(\text{Box}(C))}\Big)
  \label{eq:overlap_loss}
\end{equation}
For disjoint classes we instead use the following loss:
\begin{equation}
  \mathcal{L}_{overlap}^- = -\log\Big(1 - \frac{\textsf{Vol}(\text{Box}(C) \cap \text{Box}({D}))}{\min(\textsf{Vol}(\text{Box}(C)), \textsf{Vol}(\text{Box}(D)))}\Big)
  \label{eq:overlap_loss_neg}
\end{equation}
To avoid large flat regions in the loss landscape, for example when two boxes are completely disjoint, \cite{dasgupta_improving_2020} proposed that boxes and intersections of boxes are interpreted as Gumbel random variables. They showed that the volume of such boxes and intersections are determined by Bessel functions which can be reasonably approximated by softplus functions. In practice, this produces smooth intersections between boxes and ensures non-zero gradients, also in cases such as disjoint boxes.

Throughout this work we use $\mathcal{L}_{\sqsubseteq}$ and $\mathcal{L}_{\sqsubseteq}^-$ as placeholders for either of the inclusion and disjointness losses introduced above. To learn box embeddings the positive and negative losses are simply added together, possibly weighted differently. We present ways of doing this in more detail in Sections~\ref{sec:gnns}~and~\ref{sec:demo-molecular-function}.

\subsection*{Regularisation losses}

\cite{dhruvesh_patel_representing_2020} proposes to regularise the volume of boxes when training box embeddings, we use the implementation by \cite{chheda_box_2021} which does this by applying the L2-norm to all sides of the box,
\begin{equation}
  R = \sum_i^n\norm{Z_i - z_i}^2
  \label{eq:reg_big_box}
\end{equation}
We also found that regularising excessively small boxes can be beneficial in certain cases. This was implemented as
\begin{equation}
  R_{\text{small}} = \sum_i^n\max\big(0, \frac{1}{\norm{Z_i-z_i}} - l_0\big),
  \label{eq:reg_small_box}
\end{equation}
with $l_0$ being a threshold determining below what size the box is penalised.

\section{Material and methods}
\label{methods}

In this section, we begin by introducing the hierarchy-aware GNN framework used throughout this work. We then describe the construction of the \emph{S. cerevisiae} KG and detail how the hierarchy-aware GNN is applied to predict gene deletion fitness using this graph. Next, we present a KG embedding approach based on the same hierarchy-aware GNN architecture, trained without an explicit prediction task, to study its representational properties. Finally, we introduce a link evaluation method based on the resulting embeddings, which is used to assess the impact of adding or modifying links in the KG.


\subsection{Hierarchy-aware GNN}


\label{sec:box_gnn}

The losses presented in Section~\ref{sec:preliminaries} have primarily been used for training shallow embeddings of class hierarchies, but they can also be applied to GNNs whose outputs parameterise box embeddings. In this work, we represent KGs as TBox-style axioms by encoding class assertions, $C(a)$, as subsumption axioms, $\{a\} \sqsubseteq C$, and role assertions, $r(a,b)$, as existential restrictions, $\{a\} \sqsubseteq \exists r.\{b\}$. The resulting graph consists only of class- and nominal-level axioms of the form
\begin{align}
  &A \sqsubseteq B \label{eq:gci0} \\
  &A \sqcap B \sqsubseteq \bot \label{eq:gci1_bot}\\
  &A \sqsubseteq \exists r.B. \label{eq:gci2}
\end{align}
We refer to axioms of the form \eqref{eq:gci0} and \eqref{eq:gci1_bot} as the class hierarchy. The existential restrictions in \eqref{eq:gci2} define all edges of the graph, meaning that relations between individuals and relations between classes are represented in a uniform way within the graph.

This TBox-style graph is then used as the computational graph for the GNN: nodes correspond to classes and nominals, while directed edges correspond to axioms of the form in \eqref{eq:gci2}. Message passing therefore propagates information along logical relations in the KG, while the hierarchy axioms will provide geometric constraints on the embeddings.

\begin{figure}[!ht]
  \centering
  \includegraphics[width=0.9\textwidth]{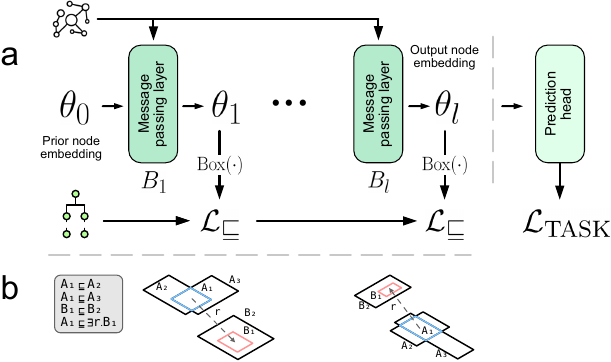}
  \caption{Overview of the Hierarchy-aware GNN. \textbf{a} shows how the output of each message passing layer, which aggregates information between neighbours in the KG, is treated as a latent variable that is converted into boxes through the box transformation in \eqref{eq:box_transform}. The boxes are trained to fulfil specified class hierarchies using the losses in (\ref{eq:dist_loss}-\ref{eq:overlap_loss_neg}), which can also be applied to the prior node embedding. The output of the GNN can be fed to some prediction head, trained by jointly minimising a task-specific loss. \textbf{b} illustrates how a box embedding is transformed throughout the GNN. Note that the relation, $r$, is drawn as an arrow in this embedding, but in the model it is represented as a message passing edge.}
  \label{fig:box_gnn}
\end{figure}

Each layer, $l\in\{1,\dots,L\}$, of a GNN learns weights $w_l$ that parameterise a function, $B_l:\mathbb{R}^{n_{l-1}}\longrightarrow\mathbb{R}^{n_l}$.
Treating the output of each layer in a GNN, $\theta_l=B_l(\theta_{l-1};w_l)$, as the latent representation of a class or nominal, box embeddings are generated via the transformation in \eqref{eq:box_transform_full} and optimised using the loss functions in (\ref{eq:dist_loss}-\ref{eq:overlap_loss_neg}). An illustration of this architecture is shown in \figureref{fig:box_gnn}.

For heterogeneous knowledge graphs spanning multiple domains whose class hierarchies can be embedded independently, we can use separate embedding spaces and hierarchy losses for each domain.

The taxonomy loss for a multilayer, possibly multidomain, GNN is calculated as the sum of the individual losses,
\begin{equation}
  \mathcal{L} = \sum_{l=0}^L\sum_{d=0}^D\sum_{(A,B)\in\mathcal{S}_d}\mathcal{L}_{\sqsubseteq}(\text{Box}(\theta_l^A), \text{Box}(\theta_l^B)),
  \label{eq:gnn_sem_loss}
\end{equation}
with $\mathcal{S}_d$ being the set of subclass axioms for domain $d$:
\begin{equation}
  \mathcal{S}_d=\{(A,B) : A\sqsubseteq B \in H^{(d)}\}
\end{equation}
and where $l$ specifies the layers of the GNN and $d$ the node domains, and $H^{(d)}$ is the class hierarchy from the ontologies describing domain $d$. $\mathcal{L}_{\sqsubseteq}$ corresponds to the loss in \eqref{eq:dist_loss} or \eqref{eq:overlap_loss}. Likewise, a negative loss can be computed to prevent the embeddings from collapsing into identical boxes for all classes. The loss is obtained either from disjointness information in the taxonomy or from randomly sampled negative examples, following the formulations in \eqref{eq:dist_loss_neg} or \eqref{eq:overlap_loss_neg}.

The approach is flexible in the sense that it is architecture agnostic and can be used either on its own for box embeddings of KGs using GNNs, or as a semantic loss together with another loss term, for example, when training prediction models. The rationale behind this approach is that class hierarchies often contain information that is not necessarily modelled in the graph edges, and can be especially useful to improve the representation of poorly connected nodes in the graph.


Training prediction models end-to-end with semantic losses, rather than first finding a separate semantic embedding of the ontology or KG, allows for extraction of the edges important for the task at hand while still adhering to the class taxonomy. The flexibility of neural networks enables this to be used for various prediction tasks, including edge, node, and graph level predictions. Depending on the prediction head it can be used for both classification and regression.

The loss function for such a model would simply combine the task specific and semantic loss, possibly along with regularisation of the parameters as
\begin{equation}
  \mathcal{L} = \mathcal{L}_{\text{TASK}}(y,\hat{y}) + \alpha (\mathcal{L}_{\sqsubseteq} + \beta \mathcal{L}_{\sqsubseteq}^-) + \lambda\norm{w}^2,
  \label{eq:combined_loss}
\end{equation}
where $\alpha$ weights the semantic loss, $\beta$ determines the impact of negative examples in the semantic loss, and $\lambda$ controls the regularisation of the network parameters.

In Section~\ref{sec:gnns}, we illustrate this approach on an edge-weight prediction (link regression) task, where the GNN output is fed into an neural network (NN) prediction head. The same architecture naturally extends to other edge-level tasks, as well as node- and graph-level tasks; the only difference lies in how the appropriate representations are selected and supplied to the NN.

To represent boxes and implement box-related operations, such as intersection and volume calculations, we use the \texttt{box-embeddings} Python package (v0.1.0)~\citep{chheda_box_2021}.

\newcommand{\specialcell}[2][c]{%
\begin{tabular}[#1]{@{}c@{}}#2\end{tabular}}

\subsection{Knowledge graph}
\label{sec:kg}

We have created a heterogeneous knowledge graph describing genes in the yeast \emph{Saccharomyces~cerevisiae} by combining facts expressed in classes and relations from multiple ontologies. We represent the graph using TBox axioms, as discussed in Section~\ref{sec:box_gnn}, by rewriting class assertions, $C(a)$, as $\{a\} \sqsubseteq C$ and role assertions, $r(a,b)$, as $\{a\} \sqsubseteq \exists r.\{b\}$. In this way, we get the same representation of asserted facts from databases as we have for terminological statements from ontologies, like GO or ChEBI. This simplifies the interface between KG, box embeddings, and GNNs.

\begin{figure}[h]
  \centering
  \includegraphics[width=\textwidth]{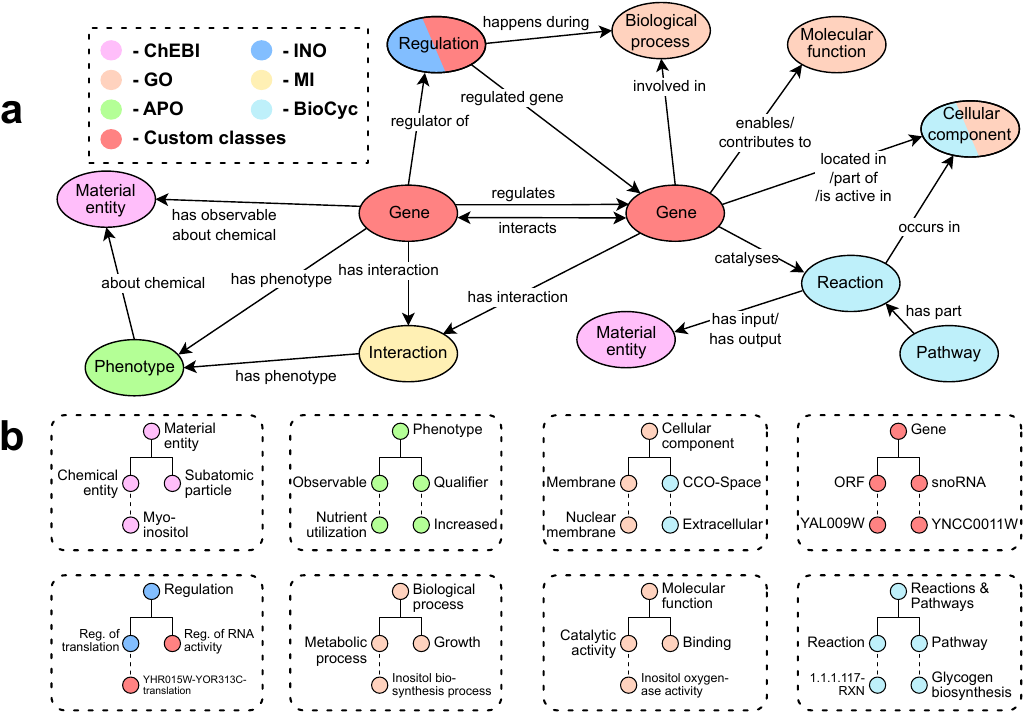}
  \vspace{-1.7em}
  \caption{An overview of the different types of classes and how they are connected in the knowledge graph is shown in \textbf{a}. The colour of the nodes specifies where the classes are defined. \textbf{b} shows examples from the hierarchies defining classes in the domains introduced in Section~\ref{sec:gnns}.}
  \vspace{-0.7em}
  \label{fig:kg}
\end{figure}

The knowledge graph is created from data in SGD, where the information is defined using terms from several different ontologies. 
A high level overview of the graph, showing how different node types are connected, can be seen in \figureref{fig:kg}a. \figureref{fig:kg}b shows examples of the hierarchies classes instantiating these nodes are represented in. Characteristics of these hierarchies are shown in \tableref{tab:kg_hierarchies}.

\begin{table}[ht]
  \centering
  \caption{Characteristics for the hierarchies representing the different domains in the KG. The ``Origin'' column indicates in which ontology or taxonomy the hierarchy is defined, $\bigstar$ indicates we have defined it, or parts of it.}
  \label{tab:kg_hierarchies}
  \begin{tabular}{|l|l|l|l|l|l|}
    \hline
    Domain & \specialcell{Number of\\classes} & \specialcell{Maximum\\depth} & \specialcell{Number of\\leaf nodes} & \specialcell{Median leaf\\node depth} & Origin \\ \hline
    Material entities & 217,381 & 24 & 200,368 & 10 & ChEBI \\ \hline
    Biological processes & 28,337 & 16 & 14,949 & 6 & GO \\ \hline
    Phenotypes & 3,949 & 9 & 3,731 & 4 & APO \\ \hline
    Molecular functions & 11,201 & 12 & 9,181 & 5 & GO \\ \hline
    Regulation & 20,563 & 7 & 20,536 & 5 & INO, $\bigstar$ \\ \hline
    Reactions & 3,101 & 8 & 2,734 & 4 & BioCyc \\ \hline
    Cellular components & 4,184 & 11 & 3,225 & 4 & GO \\ \hline
    Genes & 7,383 & 7 & 7,322 & 3 & $\bigstar$ \\ \hline
  \end{tabular}
\end{table}

The GO-annotations in SGD are naturally described by classes in the Gene Ontology and relations from the OBO Relations Ontology, which are specified in the database. Phenotypes are described using terms from APO where a phenotype is represented by an `\texttt{observable}', for example `\texttt{heat sensitivity}', and possibly a `\texttt{qualifier}', for example `\texttt{increased}'. We represent the phenotype as the subclass of the intersection of these two types of classes, and phenotypes are linked to genes using the RO relation `\texttt{has phenotype}'. Some phenotypes describe observables related to specific chemicals, in such cases the chemical class in ChEBI is linked with a custom relation, `\texttt{aboutChemical}'. To form a closer connection between genes and chemicals related to phenotypes, which proved useful for downstream tasks (see Section~\ref{sec:gnns}), a link specific to the type of observable was added between the gene and the chemical. An example of how this is implemented in description logic can be seen in \eqref{eq:dl_pheno} in Appendix~\ref{apd:dl_kg}.

Gene regulation in SGD is a directed relationship between two genes that can be positive, negative, or unspecified, and of different types, for example, regulation of protein activity or expression. In some instances, a biological process from GO specifies under which conditions the regulation occurs. We introduce custom relations describing regulation type and direction, which we use to link the two genes in the graph. When a biological process is specified we also link the genes to a gene-specific subclass of the `\texttt{regulation}' class from INO, which in turn is linked to the GO-term. The description logic implementation of such a regulation can be seen in \eqref{eq:dl_reg} in Appendix~\ref{apd:dl_kg}.

Interactions between genes are represented as undirected relationships, as the available interaction data captures symmetric associations rather than directional or causal effects between genes. These interactions may also be associated with a phenotype observed alongside the interaction. Similarly to regulation this is modelled as a link between the involved genes and a gene specific subclass of either a `\texttt{protein-protein interaction}' from INO or a `\texttt{genetic interaction}' from MI, which is linked to the phenotype.


Beyond the data from SGD we have also included information about reactions and pathways from BioCyc, which uses its own controlled vocabulary. In the graph, reactions are linked to their input and output chemicals, as well as, when specified, genes they are catalysed by and locations in the cell where they take place. We link pathways to their involved reactions, as well as to the compounds that are consumed and produced.



\subsection{Predicting gene deletion fitness}
\label{sec:gnns}



To demonstrate the usefulness of our KG, and how the method described in Section~\ref{sec:box_gnn} can be used in practice, we trained GNNs to predict phenotypic traits in \emph{S.~cerevisiae}. We use data from \cite{costanzo_global_2016} where cell growth is measured when pairs of genes are deleted (digenic deletions) from the genome. By comparing this growth to that of cells with no gene deletions, a fitness score could be determined that describes the impact of deleting the two genes. A subset of this data, grown under the same standard experimental conditions ($30^{\circ}$C), is used to train our model. This results in a dataset with 10,085,183 examples of deleted gene pairs and a corresponding fitness. Note that the genetic interaction relation from SGD describes similar phenomena, often derived from the same dataset. These relations are thus removed from the graph before training to avoid data leakage.

The prediction task can be formulated as estimating positive real-valued weights on undirected edges between gene nodes. Given a knowledge graph, $\mathcal{G} = (\mathcal{V, E, R})$, and the set of genes $\mathcal{V}_g \subset \mathcal{V}$ we aim to learn a symmetric function,
\begin{equation}
  f_\mathcal{G}: \mathcal{V}_g \times \mathcal{V}_g \rightarrow \mathbb{R}_{\geq 0},
\end{equation}
where $f_\mathcal{G}(i,j) = f_\mathcal{G}(j,i)$ predicts the fitness of the double-gene deletion $(i,j)$ based on the information encoded in $\mathcal{G}$.

We divided the classes in the KG into the eight domains in \figureref{fig:kg}b and \tableref{tab:kg_hierarchies}. This split was beneficial for performance and for stability in training (see \tableref{tab:gnn_architecture} in Appendix~\ref{apd:pred_performance}).
These splits were done manually, but generally they align well with the ontologies the classes are from, or disjoint branches in the same ontology.
The reasoning behind this is that these domains represent non-overlapping concepts, so not much is gained by representing them in the same embedding space. Doing this also allows us to reduce the dimensionality of the embedding space and vary it depending on the number of classes in the domain, thereby reducing the overall computational complexity.

After adding reverse links to enable message passing in both directions and removing infrequent edges (fewer than 1,000 occurrences), the resulting graph contains 72 distinct link types\footnote{Filtering out links with fewer than 1,000 examples removes 160 edge types, of which 118 have fewer than 100 examples and 68 have fewer than 10.} and nodes from eight different domains used for prediction. Ignoring infrequent edges was found to increase predictive performance, likely by reducing overfitting, which can be seen in \tableref{tab:ignored_edges} in Appendix~\ref{apd:pred_performance}.


Prior shallow node embeddings were trained using the $overlap$ losses in \eqref{eq:overlap_loss} and \eqref{eq:overlap_loss_neg}, representing the classes as Gumbel boxes. Large boxes were penalised using the regularisation in \eqref{eq:reg_big_box} and negative examples are generated by drawing random classes, $\bar{p}$, that are not in $\{p | c \sqsubseteq^* p\}$, where $\sqsubseteq^*$ refers to chains of the `\texttt{subClassOf}' relation, i.e., negative examples are not in the set of all ancestors to $c$. Parameters used to train the box embeddings and the dimensions of the different domains are reported in Appendix~\ref{apd:box_hyper}.

\begin{figure}[!ht]
  \centering
  \includegraphics[width=0.9\textwidth]{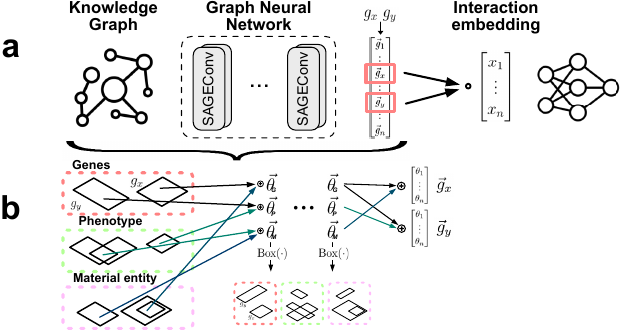}
  \vspace{-0.7em}
  \caption{An overview of the system predicting the fitness when deleting pairs of genes is shown in \textbf{a}. A GNN using GraphSAGE message passing layers, acting on the KG from Section~\ref{sec:kg}, generates node embeddings. The embeddings of pairs of genes are combined through element-wise multiplication and fed to a NN predicting the fitness of the gene deletion. \textbf{b} shows how classes in the different domains are represented by boxes and how information is aggregated in the GNN, as well as how the node embeddings throughout the network are interpreted as boxes using the box transformation in \eqref{eq:box_transform}. These boxes are optimised to represent the class hierarchies from the underlying ontologies. Arrows represent learnable GraphSAGE modules, different for each \texttt{source domain-edge-target domain} type.
  }
  \label{fig:architecture}
\end{figure}

For predicting the gene-pair fitness we use a heterogeneous GNN together with a fully connected neural network; an overview of the architecture can be seen in \figureref{fig:architecture}a.
The GNN used is based on the max-aggregated GraphSAGE embedding algorithm~\citep{hamilton_inductive_2017} briefly introduced by in Section~\ref{sec:preliminaries}. In our heterogeneous setting, each source-edge-target type has its own SAGEConv-module as proposed by \cite{schlichtkrull_modeling_2018}, whose outputs are combined using mean aggregation to create the node embeddings from each layer. We found it beneficial to adjust the dimensionality of the message-passing modules based on the type of source-edge-target triple. Domains with a high degree of incoming connectivity, such as `Material entities' or `Genes', are assigned higher dimensional feature spaces. This can be interpreted as increasing the expressivity for domains with high degree connectivity and the performance gain can be observed observed in \tableref{tab:gnn_dimensions} in Appendix~\ref{apd:pred_performance}.  The resulting class embeddings, generated by the GNN, capture aggregated neighbourhood information. By applying box-losses as introduced in Section~\ref{sec:box_gnn}, the training will also aim to represent the class hierarchy as box embeddings. \figureref{fig:architecture}b illustrates how information is propagated from the initial box embeddings across different domains to the gene embeddings.

Because the task operates at the edge level, the embeddings of the deleted genes must be combined prior to being fed into the regressor network. This can be done in a number of ways, in this work we have considered four ways of doing this: element-wise product, vector-valued bilinear symmetric transformation, box intersection, and concatenation, defined in \tableref{tab:gene_combination}. Among these, the product, bilinear, and intersection methods are symmetric, and the bilinear transformation is the only learnable combination method. As can be seen in \tableref{tab:gene_combine_performance} in Appendix~\ref{apd:pred_performance}, we found the element-wise product to perform the best. The node-pair representation is fed to a fully connected neural network outputting a real valued prediction.

\begin{table}
  \centering
  \caption{Node feature combinations for edge-level prediction task.}
  \label{tab:gene_combination}
  \begin{tabular}{|l|l|c|c|} \hline
    & Formulation & Symmetric & Learnable \\ \hline
    Product & $x_1 \odot x_2$ & $\checkmark$ & \\ \hline
    Bilinear & $x_1 \odot Wx_2 + x_2 \odot Wx_1$ & $\checkmark$ & $\checkmark$ \\ \hline
    Intersection & $\text{Box}(x_1) \cap \text{Box}(x_2)$ & $\checkmark$ & \\ \hline
    Concatenation & $\mathtt{concat}(x_1, x_2)$ & & \\ \hline
  \end{tabular}
\end{table}


We train the model by minimising, using the Adam optimiser, the following loss function,
\begin{equation}
  \mathcal{L} = \mathcal{L}_{\text{MSE}}(y,\hat{y}) + \alpha (\mathcal{L}_{\sqsubseteq} + \beta \mathcal{L}_{\sqsubseteq}^-) + \lambda\norm{w}^2
  \label{eq:sem_loss}
\end{equation}
$\mathcal{L}_{\text{MSE}}$ denotes the mean squared errors of the fitness predictions and $\mathcal{L}_{\sqsubseteq}$ and $\mathcal{L}_{\sqsubseteq}^-$ are defined in \eqref{eq:gnn_sem_loss}. $\alpha$ and $\beta$ are weights determining the impact of the semantic loss, measuring how well the box embeddings represents the class hierarchies. $\lambda$ controls regularisation of the parameters in the network. 

The models are trained and evaluated using 10-fold cross validation where the data split is based on the genes. Any gene pairs that include genes from both the training and validation sets are discarded. This ensures that no pairs involving validation-set genes are seen during training, so the learned representations of genes in the training set do not influence the predictions being evaluated.

Hyperparameters, including learning rate, regularisation ($\lambda$), the depth and width of the fully connected neural network, the depth of the GNN, and embedding dimensions throughout the GNN, are tuned using Bayesian optimisation. Tuning is performed on a separate data split from the one evaluated in Section~\ref{sec:inter_preds}. This tuning is done for a model using box embeddings as prior node representations, but without semantic loss during training, the same parameters are then used for all evaluated models. The used hyperparameters are reported in Appendix~\ref{apd:model_hyper}.

\subsection{Hypothesis generation}
\label{sec:hypo_gen}
Because the predictions are derived from a knowledge graph in which each edge encodes domain-relevant semantics, they can be exploited to identify patterns among the most informative relationships in the graph. Furthermore, because the predicted measures can be interpreted as arising from interactions between the two deleted genes, we examine which gene-associated traits are jointly influential in driving the model's predictions.

The procedure for identifying such interactions is given in Algorithm~\ref{alg:hypothesis_gen}. Using gradient-based post-hoc interpretability methods, we assign importance scores to all nodes connected to the genes involved in each deletion. For each pair of gene-associated nodes, we compute an interaction score as the product of their individual importance values, and then aggregate these scores across deletions to obtain a set of globally important interacting traits. Note that this method does not rely on hierarchy-aware GNN embeddings.

\begin{algorithm2e}[H]
  \LinesNumbered
  Model \# trained prediction model \\
  Explainer(Model) \# importance attribution algorithm \\
  importances $\leftarrow \{\}$ \\
  \ForEach{$\{g_1,g_2\}\in \{\text{Deleted genes}\}$}{
    ($links\_g_1, links\_g_2$) $\leftarrow$ Explainer($g_1, g_2$) \# importance scores for ($s, p$) pairs from triples where the $object$ is $g_1$ and $g_2$ respectively \\
     \ForEach{$l_1 \in links\_g_1$}{
        \ForEach{$l_2 \in links\_g_2$}{
          importances[$l_1l_2$] $\leftarrow$ importances[$l_1l_2$] + $l_1 * l_2$
      }
    }
  }
  \caption{Edge pair importance attribution algorithm}
  \label{alg:hypothesis_gen}
\end{algorithm2e}

We used the $input \times gradient$ method~\citep{shrikumar_not_2017} to assign importance scores, implemented in Captum~\citep{kokhlikyan2020captum}, but this approach is not specific to any attribution method.


\subsection{Learning GNN box embeddings without a prediction task}\label{sec:demo-molecular-function}

To study the effect that semantic losses have on embedding representations, and to demonstrate how they can be used to train box embeddings in the absence of a prediction task, we use the same KG as above with some modifications. Following the same methodology as Section~\ref{sec:kg} we rewrite role and class assertions as TBox axioms. `\texttt{subClassOf}' relations are used as the positive examples for~$\mathcal{L}_{\sqsubseteq}$, and negative examples for~$\mathcal{L}_{\sqsubseteq}^-$ are taken from disjointness axioms from two sources. Firstly, we create additional TBox statements for disjointness between the three subclasses in the molecular function domain, and between each of these classes and the subclasses of the other two. Secondly, for randomly drawn pairs to distinguish between individual classes in the graph. 

In the same way as the models described in Section~\ref{sec:gnns}, the GNN is constructed from SAGEConv modules for each edge type. For the purposes of demonstration, we learn embeddings in two dimensions so they can easily be visualised. We simultaneously train initial box embeddings (randomly initialised) and a GNN by minimising, again using the Adam optimiser, the loss function:
\begin{equation}
  \mathcal{L} = \mathcal{L}_{\sqsubseteq} + \beta \mathcal{L}_{\sqsubseteq}^- + \lambda_{s} R_{\text{small}} + \lambda\norm{w}^2,
  \label{eq:demo_box_loss}
\end{equation}
where $R$ is the regularisation loss from \eqref{eq:reg_small_box}, penalising small boxes with an $l_0$ of 1, and $\beta$, $\lambda$ and $\lambda_{s}$ are weights determining the impact of the negative semantic loss, weight regularisation loss, and small box regularisation loss respectively. Note the absence of the mean squared error term present in the prediction task above. The regularisation term was included as disjointness tended to make boxes extremely small along one or more dimensions during training rather than move position in the space. The negative semantic loss is decomposed into
\begin{equation}
  \mathcal{L}_{\sqsubseteq}^- = \mathcal{L}_{\sqsubseteq\text{data}}^- + \gamma\mathcal{L}_{\sqsubseteq\text{random}}^-,
  \label{eq:demo_box_negative_loss}
\end{equation}
which enables us to tune the contribution of the randomly selected disjointness axioms.
For each loss type ($distance$ or $overlap$) we separately tuned the hyperparameters, and the used values are reported in \tableref{tab:family_box_hyper} in Appendix~\ref{apd:demo_hyper}.



\subsection{Link evaluation}\label{sec:link_eval_method}


A potential application for the semantic losses defined above, in addition to the training of box embedding models for quantitative prediction tasks, is to rank proposed revisions to a knowledge graph based on the resultant changes to embeddings and losses. Distance based approaches to link prediction have been attempted before, for example with HAKE~\citep{zhang_learning_2020} where they replaced either subject or object in existing triples. But the use of a GNN in our method means we can theoretically assess completely unseen additions to the graph by evaluating their global effect. We test this by adding single edges to the graph according to the following scheme.

Say that for a given ontology we construct a graph $\mathcal{G}=(V,E)$, where each edge vertex $v\in V$ is a class in the ontology and each edge $e\in E$ represents a role assertion. Following the methodology outlined above, we use $\mathcal{G}$ as the basis for a GNN, and train box embeddings and the weights of the GNN using the semantic loss. Introducing new role assertions to the graph results in graph $\tilde{\mathcal{G}}$, and passing the prior box embeddings through the GNN using these additional edges will change the final box embeddings. We calculate the distance between the original learned box embeddings and those after the changes to the graph, giving us a measure of the change to the embeddings from the graph revision. This process is described in Algorithm~\ref{alg:link_eval}.


\begin{algorithm2e}[H]
  \LinesNumbered
  Train embedding parameters $\theta_l$ on $\mathcal{G}_{\text{train}}$ \\
  $\delta \leftarrow \emptyset$ \\
  \ForEach{$e\in E_{\text{test}}$}{
  $\tilde{\mathcal{G}} \leftarrow \mathcal{G}_{\text{train}} \cup \{e\}$ \\ 
  $B \leftarrow \text{Box}(\text{GNN}_\theta(\mathcal{G}_{\text{train}}))$ \\
  $\tilde{B} \leftarrow \text{Box}(\text{GNN}_\theta(\tilde{\mathcal{G}}))$ \\
  $\delta \leftarrow \delta \cup (e,\langle B, \tilde{B}\rangle)$ \# distance between the generated box embeddings \\ 
  }
  Sort $\delta$ to get ranked revisions
  \caption{Link evaluation algorithm}
  \label{alg:link_eval}
\end{algorithm2e}



To evaluate this proposed method, we split the edges in the KG into training and test data using a 80:20 training and test split, stratified by relation type. This results in a training graph $\mathcal{G}_{\text{train}}=(V,E_{\text{train}})$ and a test graph $\mathcal{G}_{\text{test}}=(V,E_{\text{test}})$. The embeddings and GNN are trained as per Section~\ref{sec:demo-molecular-function}. We run Algorithm~\ref{alg:link_eval}, going through each edge in the test data. We also perform the same steps with randomly generated edges with source and target drawn from the same classes as the test edge, and with completely randomly drawn source and target nodes. The distance metric used is defined in \eqref{eq:box_distance}.


\section{Results}
\label{sec:results}

\subsection{Gene deletion fitness prediction}
\label{sec:inter_preds}

In Table~\ref{tab:results} we present the coefficient of determination ($R^2$) for different versions of the model described in Section~\ref{sec:gnns}. We evaluate a model without any information from class hierarchies, a model where the hierarchical information is introduced as \texttt{subClassOf}-links in the KG, a model with the prior node embeddings in box form, and models using both prior node embeddings and the semantic loss in \eqref{eq:sem_loss}. Both the $overlap$ and $distance$ version of the losses are evaluated. The model not using any hierarchy information ($c$ in \tableref{tab:results}) and the model where the hierarchies are represented by links in the graph (($d$ in \tableref{tab:results})), learns shallow embeddings specifically for this task to represent the nodes in the KG. For the model only using prior box embeddings ($e$ in \tableref{tab:results}), these are not modified during training.
For the two models with the semantic loss ($f$ and $g$), we apply it to all domains except the one embedding the genes. The gene hierarchy builds on a rudimentary SGD gene categorisation, offering very little information, with over 90\% of genes falling into the same category. Hence we deem the hierarchy in this domain uninformative.

The GNN-based models are compared to Light Gradient Boosting Machines (LightGBMs)~\citep{ke_lightgbm_2017}. We considered the instantiation of the phenotype information from the KG as gene representation. The phenotypes describe observable characteristics of the genes and are the part of the KG we expect to be most informative for this task (further support for this is seen when considering feature importances for the GNN, mentioned in Section~\ref{sec:bio_exp}, which are dominated by phenotypes). The instantiation of the phenotypes is sparse, with 2680 features.

We also considered a ComplEx~\citep{trouillon_complex_2016} embedding (64 dimensions) of the KG as an alternative gene representation. In contrast to the phenotype instantiation, which relies on a hand-selected subset of biologically relevant relations, ComplEx provides a dense, low-dimensional embedding learned from the full multi-relational structure of the KG. This allows LightGBM to exploit global relational patterns that are not explicitly encoded in the phenotype feature vectors. We also evaluated ComplEx embedding with an MLP prediction head, as well as a Box$^2$EL-based predictor. They did not achieve the same predictive performance, and are reported in \tableref{tab:baselines} in Appendix~\ref{apd:pred_performance}.

\begin{table}[h]
  \centering
  \caption{Results from 10-fold cross-validation of digenic deletion fitness. The GNNs in $c$ and $d$ learn task-specific shallow embeddings, on non-box form, as the prior node representations. In $d$ the hierarchical information is introduced as links in the KG. The other three GNNs ($e$-$g$) uses pre-trained box embeddings and the semantic loss in \eqref{eq:sem_loss} is applied to two of them ($f$ and $g$). All GNN models share the same architecture. The instantiation model ($a$) uses a sparse feature matrix with non-zero entries for phenotype annotations from the KG. The ComplEx based model ($b$) first embeds the KG in 64 dimensions and predicts the growth from this. Significant pairwise differences are indicated by $\uparrow$ and $\downarrow$ ($p$$<$0.05, paired, one-sided t-test).}
  \label{tab:results}
  \vspace{-0.7em}
  \begin{tabular}{|l|l|l|l|l|l|l|l|l|l|l|}
    \hline
    & Description & Mean $R^2$ & SD & $a$ & $b$ & $c$ & $d$ & $e$ & $f$ & $g$ \\ \hline
    $a$ & ComplEx + LightGBM & 0.191 & 0.039 & - & & $\downarrow$ & $\downarrow$ & $\downarrow$ & $\downarrow$ & $\downarrow$ \\ \hline
    $b$ & Instantiations + LightGBM & 0.211 & 0.022 & & - & $\downarrow$ & $\downarrow$ & $\downarrow$ & $\downarrow$ & $\downarrow$ \\ \hline
    $c$ & GNN without box embeddings & 0.348 & 0.050 & $\uparrow$ & $\uparrow$ & - & & & $\downarrow$ & $\downarrow$ \\ \hline
    $d$ & GNN with \texttt{subClassOf}-links in KG & 0.350 & 0.049 & $\uparrow$ & $\uparrow$ & & - & & $\downarrow$ & $\downarrow$ \\ \hline
    $e$ & GNN with prior box embeddings & 0.360 & 0.043 & $\uparrow$ & $\uparrow$ & & & - & & $\downarrow$ \\ \hline
    $f$ & GNN + $\mathcal{L}_{overlap}$ & 0.368 & 0.038 & $\uparrow$ & $\uparrow$ & $\uparrow$ & $\uparrow$ &  $ $ & - & $ $ \\ \hline
    $g$ & GNN + $\mathcal{L}_{distance}$ & \textbf{0.377} & 0.046 & $\uparrow$ & $\uparrow$ & $\uparrow$ & $\uparrow$ & $\uparrow$ & $ $ & - \\ \hline
  \end{tabular}
\end{table}

From the results it is clear that the GNN generates gene embeddings which can be used for predicting this fitness to a reasonable degree, given the amount of noise typically present in biological measurements~\citep{li_performance_2021}. Using the box embeddings to represent classes rather than learning them from scratch results in a significant ($p$$<$0.05, paired t-test) improvement. Enforcing the hierarchical class structure through the semantic loss throughout the model improves the results further, the model trained with the $distance$-loss performs significantly ($p$$<$0.05, paired t-test) better than the models not using the semantic loss. Interestingly, the box representations also outperforms explicitly adding the \texttt{subClassOf}-information to the KG, suggesting the inclusion of boxes is a useful inductive bias for these representation. The instantiated phenotype information seems to be somewhat useful for prediction, but is not as informative as the full KG. We also see that the ComplEx KG embedding captures information useful for this task, however not as informative as the task specific embedding found by the GNNs.

\begin{figure}[h]
\centering
{%
  \subfigure[Double deletion, $R^2=0.360$]{\label{fig:double_parity}%
  \includegraphics[height=0.426\linewidth]{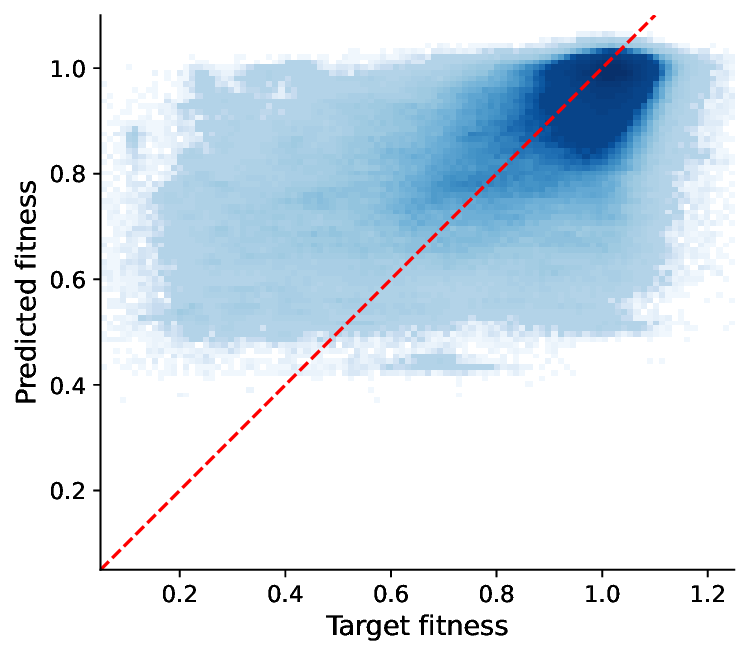}}%
\quad
\subfigure[Triple deletion, $R^2=0.380$]{\label{fig:triple_parity}%
  \includegraphics[height=0.426\linewidth]{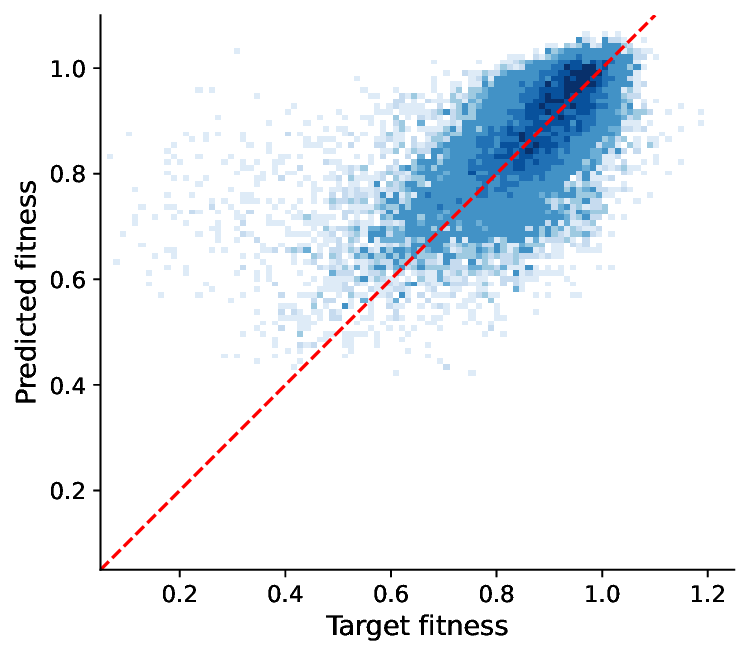}}
  \\
  \subfigure[{\centering
Double deletion with semantic loss, $R^2 = 0.377$}]%
  {\label{fig:double_parity_sem}\includegraphics[height=0.426\linewidth]{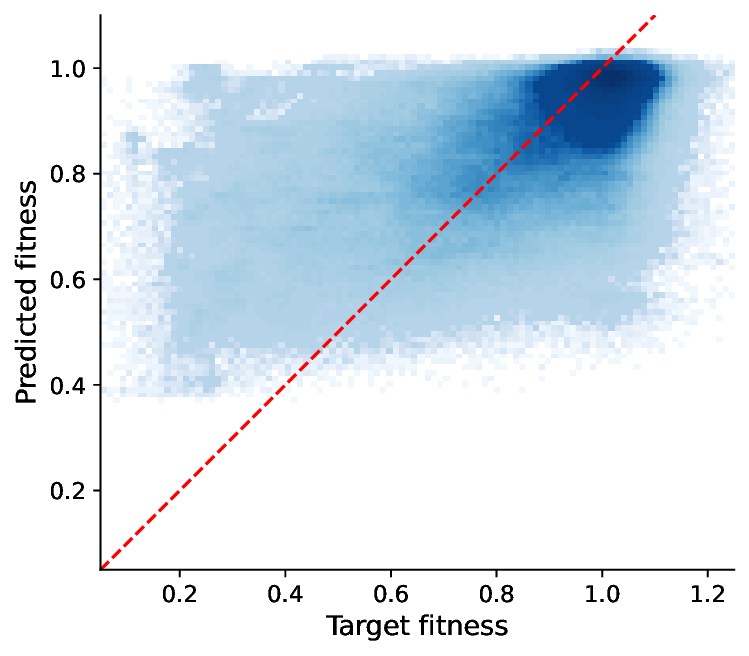}}%
\quad
\subfigure[{\centering Triple deletion with semantic loss, $R^2=0.415$}]{\label{fig:triple_parity_sem}%
  \includegraphics[height=0.426\linewidth]{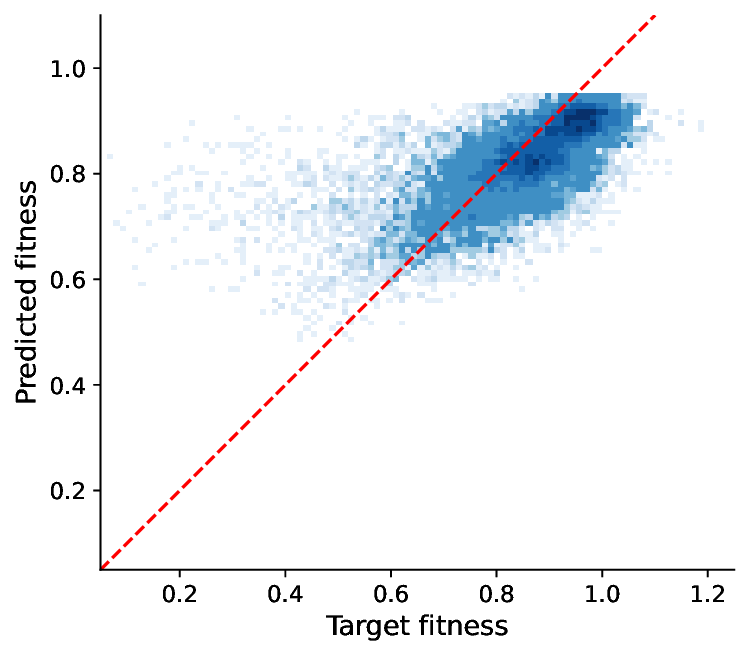}}
}
\vspace{-0.9em}
\caption{Parity plots for double, $(a)$ and $(c)$, and triple, $(b)$ and $(d)$, gene deletions. $(a)$ and $(b)$ shows the parity plot for the model using box embeddings as prior node representations only, while $(c)$ and $(d)$ shows the predictions from a model also trained with the $distance$-based semantic losses, $\mathcal{L}_{distance}$. For the double deletion, the predictions from all validation sets in the cross validation are shown.}
\label{fig:parity}
\end{figure}

The parity plots for the predictions are shown in \figureref{fig:double_parity}~and~\ref{fig:double_parity_sem}. From this we can see that most double gene deletions do not have major impact on the fitness. We can also see a clear shrinkage effect where the model mispredicts extreme values, especially deletions with low fitness are overestimated. Comparing the predictions from the model trained with the semantic loss we can see that they in general are rather similar, but that the semantic loss model seems to have fewer large underestimations.

\begin{figure}[h]
  \centering
  \includegraphics[width=\textwidth]{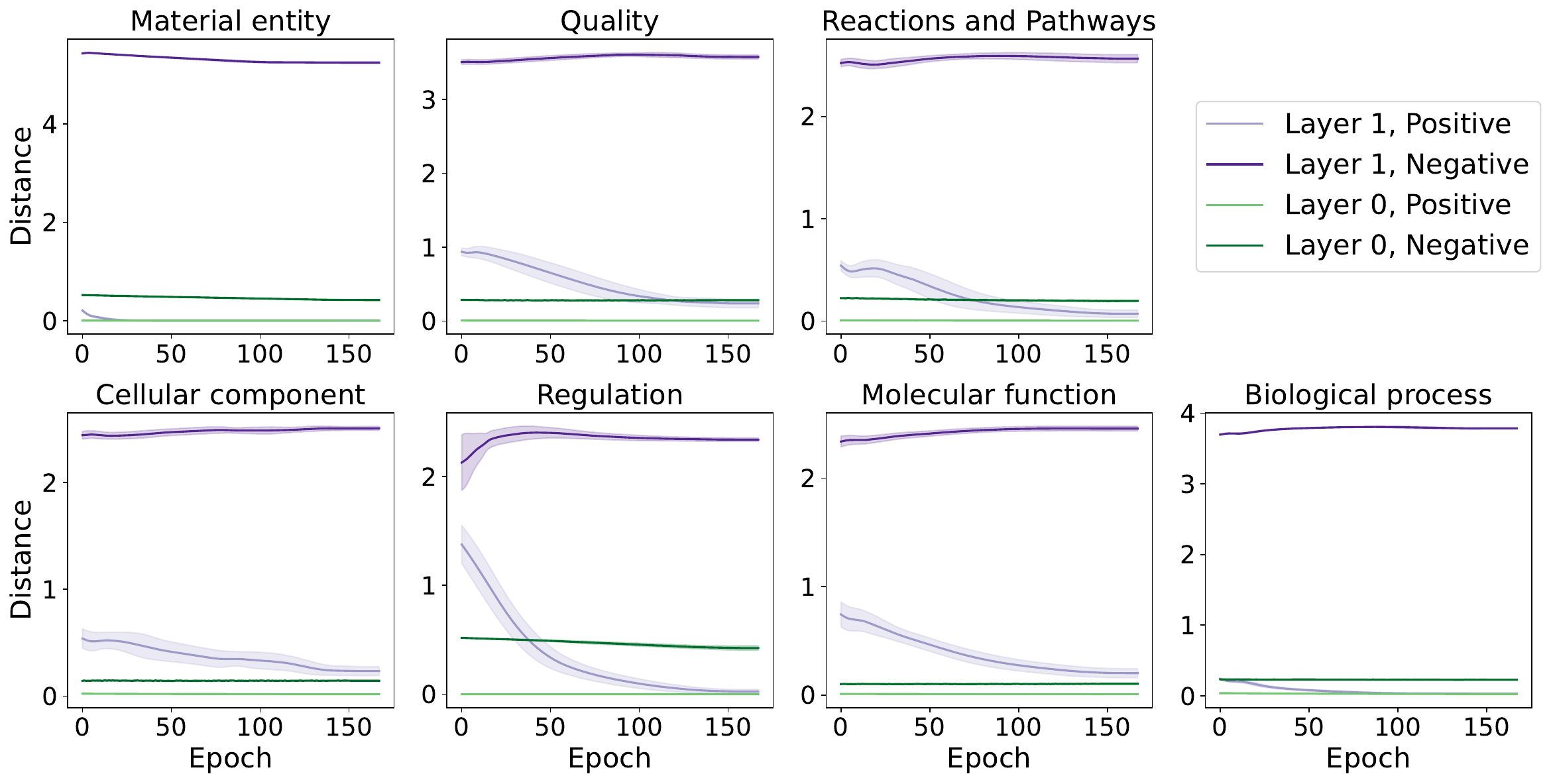}
  \caption{Average $\mathcal{L}_{distance}^{}$ and $\mathcal{L}_{distance}^-$ losses per class for the different domains in the KG, during training of the best performing model ($f$) in \tableref{tab:results}. The line is the average loss across the 10 folds and the shaded area shows $\pm$ one standard deviation. Note that this loss is not applied to the gene-domain, since its class hierarchy is not deemed to be informative.}
  \label{fig:sem_losses}
\end{figure}

\figureref{fig:sem_losses} show the semantic losses in the different domains, introduced in \figureref{fig:kg}b and \tableref{tab:kg_hierarchies}, for the best performing model, using $\mathcal{L}_{distance}$. A similar pattern is observed for all domains where both the positive loss for the first layer (the pretrained box embeddings) and the negative loss for the second layer is low and fairly constant. The positive losses for the second layers decreases across all domains throughout training, while the negative loss for the first layer does not change much and is substantially higher than the others. This suggests that, even though they result in a significant improvement in prediction performance, the initial box embeddings have a lot of overlap between classes. On the other hand, the embeddings generated by the GNN discriminate very well between classes, already at the first epoch and the hierarchical structure is learnt throughout training.

To evaluate our model on a slightly modified version of the original task we used data from \cite{kuzmin_systematic_2018}, who performed a study similar to the one used for training our models, but focused on trigenic deletion fitness. This dataset comprises a total of 15,095 triple deletion datapoints. For this we use one model trained on the full dataset from \cite{costanzo_global_2016}, but instead perform the element-wise product between the three involved genes. Notably we achieve an $R^2$ of 0.380 for a model using box embeddings as prior node representations, and 0.415 for a model using the same prior node embeddings, but trained with the $distance$-based semantic loss. These values are slightly higher than the average performance observed in the cross-validation of digenic deletions. \figureref{fig:triple_parity}~and~\ref{fig:triple_parity_sem}, shows parity plots for these predictions. Again, the prediction patterns for the two models look similar, but the model trained with the semantic loss does not predict as high fitness. An important note on this experiment is that, unlike the double deletion experiment, the individual genes making up the triple deletions are now seen as parts of double deletion examples in training.

\subsection{Hypothesis generation and experimental evaluation}
\label{sec:bio_exp}

In this section, we present a case study evaluating the interaction-discovery procedure described in Section~\ref{sec:hypo_gen}, identifying candidate trait interactions for experimental testing. To focus on patterns corresponding to viable experiments for our laboratory setup, we filtered for edges related to nutrient utilisation phenotypes. A more detailed description of the filtering process can be found in Appendix~\ref{apd:exp_filtering}. The top ten most important edge pairs are shown in \figureref{fig:feat_imp} and detailed in \tableref{tab:apo_nutrient_classes} and \ref{tab:apd_allowed_edge} in Appendix~\ref{apd:top_edges}. The highest-weighted, safely testable pair was selected and highlighted in red in \figureref{fig:feat_imp}, linking one of the involved genes to inositol (vitamin B8) utilisation and the other to NaCl stress resistance, suggesting a potential interaction between these traits.


\begin{figure}[h]
  \setcounter{subfigure}{0}
  \centering
    {%
    \subfigure[Feature importances]{\label{fig:feat_imp}%
    \includegraphics[height=0.335\linewidth]{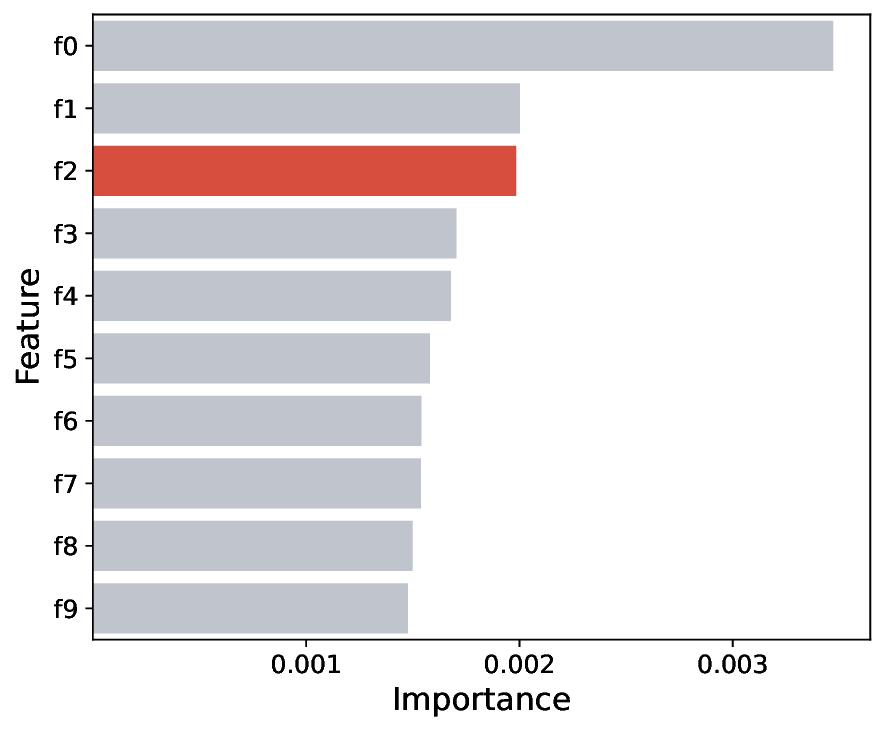}}%
  \quad
    \subfigure[AUC growth boxplots]{\label{fig:fowth_box}%
    \includegraphics[height=0.335\linewidth]{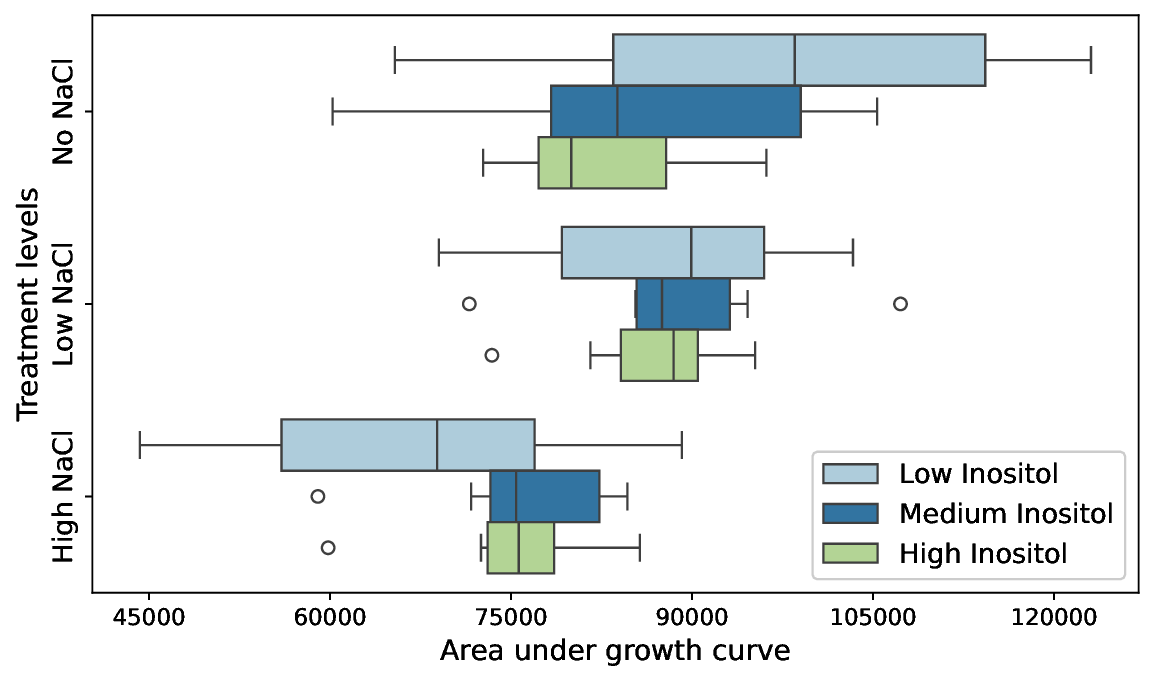}}
  }
  \vspace{-1.9em}
  \caption{An overview of the selection and results of the experiment we performed. ($a$) shows the highest ranked importances of edge-pairs and the pair selected for the experiment, nutrient utilisation of inositol and stress resistance to NaCl, is highlighted in red. \emph{f0} and \emph{f1}, which have a higher assigned weight, are discarded due to safety and lab constraints as it involves the chemical bleomycin. ($b$) Box plot showing the distribution of AUC for all of the experimental conditions tested. Inositol supplementation significantly impacts growth dynamics in high doses ($p$ $<$ 0.05). NaCl stress changes the impact of inositol in a dose dependent manner, suggesting an interactive effect ($p$ $<$ 0.05).}
  \label{fig:exp_results}
\end{figure}

To experimentally test this hypothesis, a perturbation experiment was performed in an automated laboratory cell~\citep{williams_cheaper_2015}, in which inositol and NaCl was supplied in a range of concentrations, details about the experimental design and cultivation methods can be found in Appendix~\ref{apd:cultivation}. An $\Delta$\emph{ino1} mutant (INOsitol requiring) was used for all subsequent experiments, as it is unable to synthesise inositol on its own, ensuring that any intracellular accumulation was acquired only through transport from the media. The growth dynamics of the cells in the different experimental conditions were summarised with the area under curve (AUC) of the growth curves, providing a single-valued measure of the biomass accumulation over the course of the experiment. The full growth dynamics can be seen in \figureref{fig:growth_curves} in Appendix~\ref{apd:stats} and summarising boxplots are shown in \figureref{fig:fowth_box}. Statistical testing for interaction effects was done with a Gaussian generalised linear model (GLM), further details can be found in Appendix~\ref{apd:stats}.

These empirical results, seen in \figureref{fig:fowth_box} and \tableref{tab:stats_results} in Appendix~\ref{apd:stats}, indicate a significant interaction between inositol supplementation and induced NaCl stress, verifying that the proposed edge-interactions are consistent with experimental data. Specifically, supplementing with inositol rescued cells from NaCl-induced stress, indicating that inositol availability enhances their ability to withstand salt stress. Inositol has previously been implicated in biosynthesis and integrity of cell membranes~\citep{culbertson_inositol-requiring_1975}. Since NaCl can disrupt osmotic balance, enhanced membrane stability is likely to have a protective effect for the cells.

\subsection{Demonstration of embeddings for the molecular function domain}
\label{sec:cell-comp-demo}

\begin{figure}[ht]
    \centering
    \subfigure[Distance Loss - Pre GNN Embeddings]{
        \includegraphics[width=0.45\linewidth]{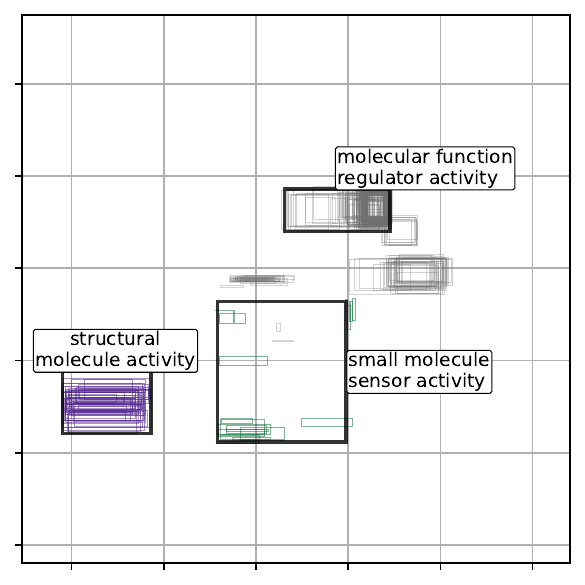}
        \label{fig:pre-gnn-distance}
    }
    \subfigure[Distance Loss - Final Embeddings]{
        \includegraphics[width=0.45\linewidth]{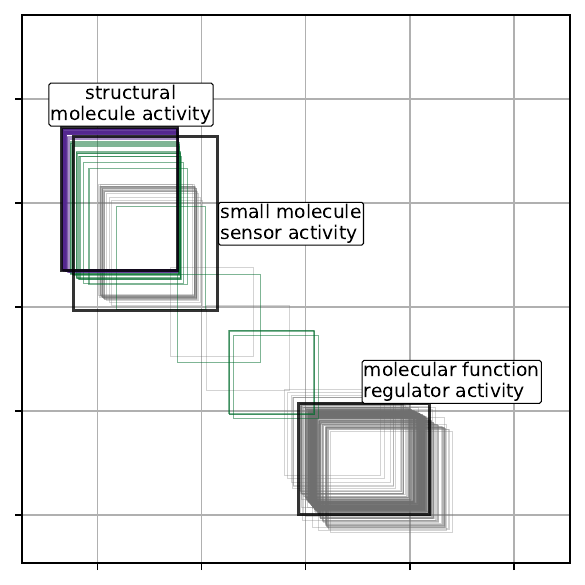}
        \label{fig:final-boxes-distance}
    }
    \\
    \subfigure[Overlap Loss - Pre GNN Embeddings]{
        \includegraphics[width=0.45\linewidth]{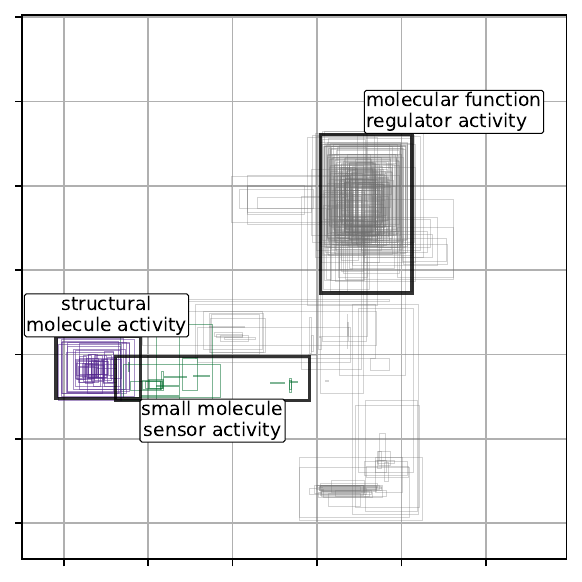}
        \label{fig:pre-gnn-overlap}
    }
    \subfigure[Overlap Loss - Final Embeddings]{
        \includegraphics[width=0.45\linewidth]{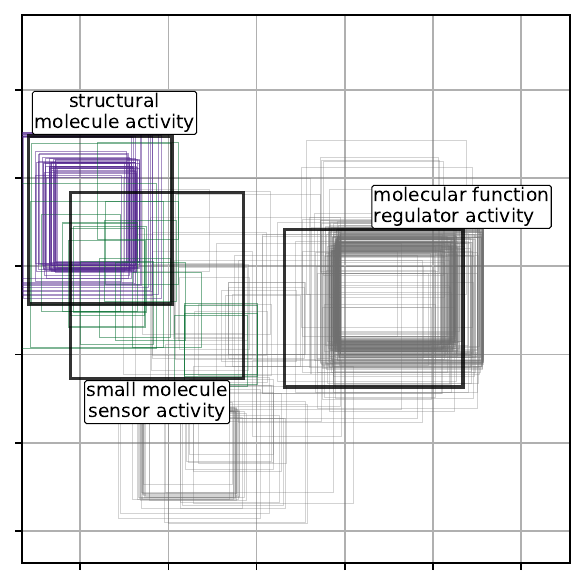}
        \label{fig:final-boxes-overlap}
    }
    \fbox{%
      \parbox{0.50\linewidth}{%
        \footnotesize
        \raggedright
        \textcolor{molfuncreg}{\rule{1.2em}{1.2ex}}~--~\texttt{molecular function regulator activity} \\
        \textcolor{strucmol}{\rule{1.2em}{1.2ex}}~--~\texttt{structural molecule activity} \\
        \textcolor{smallmolsens}{\rule{1.2em}{1.2ex}}~--~\texttt{small molecule sensor activity}
      }
    }
    \vspace{0.0em}
    \caption{Learned box embeddings in two dimensions for the molecular function domain. \ref{fig:pre-gnn-distance}~and~\ref{fig:final-boxes-distance} box embeddings prior to input into GNN; \ref{fig:final-boxes-distance}~and~\ref{fig:final-boxes-overlap} show final embeddings for $distance$ and $overlap$ loss respectively.}
    \label{fig:trained-boxes-comparison}
\end{figure}

For the knowledge graph constructed according to the method in Section~\ref{sec:demo-molecular-function}, after the hyperparameter search for $distance$-loss and $overlap$-loss, we constructed final box embeddings in two dimensions using a GNN with one message passing layer. In Figure~\ref{fig:trained-boxes-comparison} we plot box embeddings for each loss, before input into the GNN and then the final embeddings, for a subset of classes in the molecular function domain. We see clearly that the GNN is performing a transformation of the prior embeddings.
Furthermore, both losses result in learned embeddings that begin to capture semantic concepts from the KG, in particular that `\texttt{structural molecule activity}' and `\texttt{molecular function regulator activity}' are disjoint, and that each of these is disjoint from `\texttt{small molecule sensor activity}'.
Both losses also responded to the randomly drawn negative loss contributions ($\mathcal{L}_{\sqsubseteq\text{random}}^-$) in attempting to separate individuals, though this is more evident with the $overlap$-loss. These semantic constraints are more evidently respected in the embeddings before the message passing in the GNN. This is perhaps understandable, as the message passing includes no additional information about the hierarchy of the ontology, but does include information from the relations in the graph.
Also, in this KG, the concepts in the molecular function domain have relations only to concepts in the gene domain (see Figure~\ref{fig:kg}). The hierarchy in the gene domain, constructed from the gene categorisation in SGD, is fairly flat. The majority of concepts in this domain are immediate subclasses of `\texttt{ORF}'.


Aside from this observation, comparing Figures~\ref{fig:pre-gnn-distance}~and~\ref{fig:final-boxes-distance}, we see that prior to being passed through the GNN, the boxes corresponding to different `\texttt{structural molecule activity}', `\texttt{molecular function regulator activity}', and `\texttt{small molecule sensor activity}' subclasses are of quite varied shape and size, and somewhat spread through the embedding space. Some clustering is apparent, which corresponds to structure in the hierarchy. The loss from randomly drawn disjointness axioms keeps them somewhat separated. Having passed through the GNN, the boxes take on dramatically different shapes and relative positions. The semantic loss of these embeddings is low on average, but subclasses of the same superclass have now very similar embeddings, and there are some clear violations of the hierarchy.
By contrast, comparing Figures~\ref{fig:pre-gnn-overlap}~and~\ref{fig:final-boxes-overlap}, we see that the shapes of the boxes trained with $overlap$ losses are often longer and thinner, minimising the volume of each intersection.
And the final embeddings do not differ from the prior embeddings to the same extent as with $distance$ loss. The overall position of the parent classes is broadly the same, with some minor translations.
With both losses, the box volume increased after passing through the GNN, and the difference between height and width decreased.

\subsection{Evaluation of graph revisions}\label{sec:link_eval_results}

\begin{figure}[h]
    \centering
    \includegraphics[width=\textwidth]{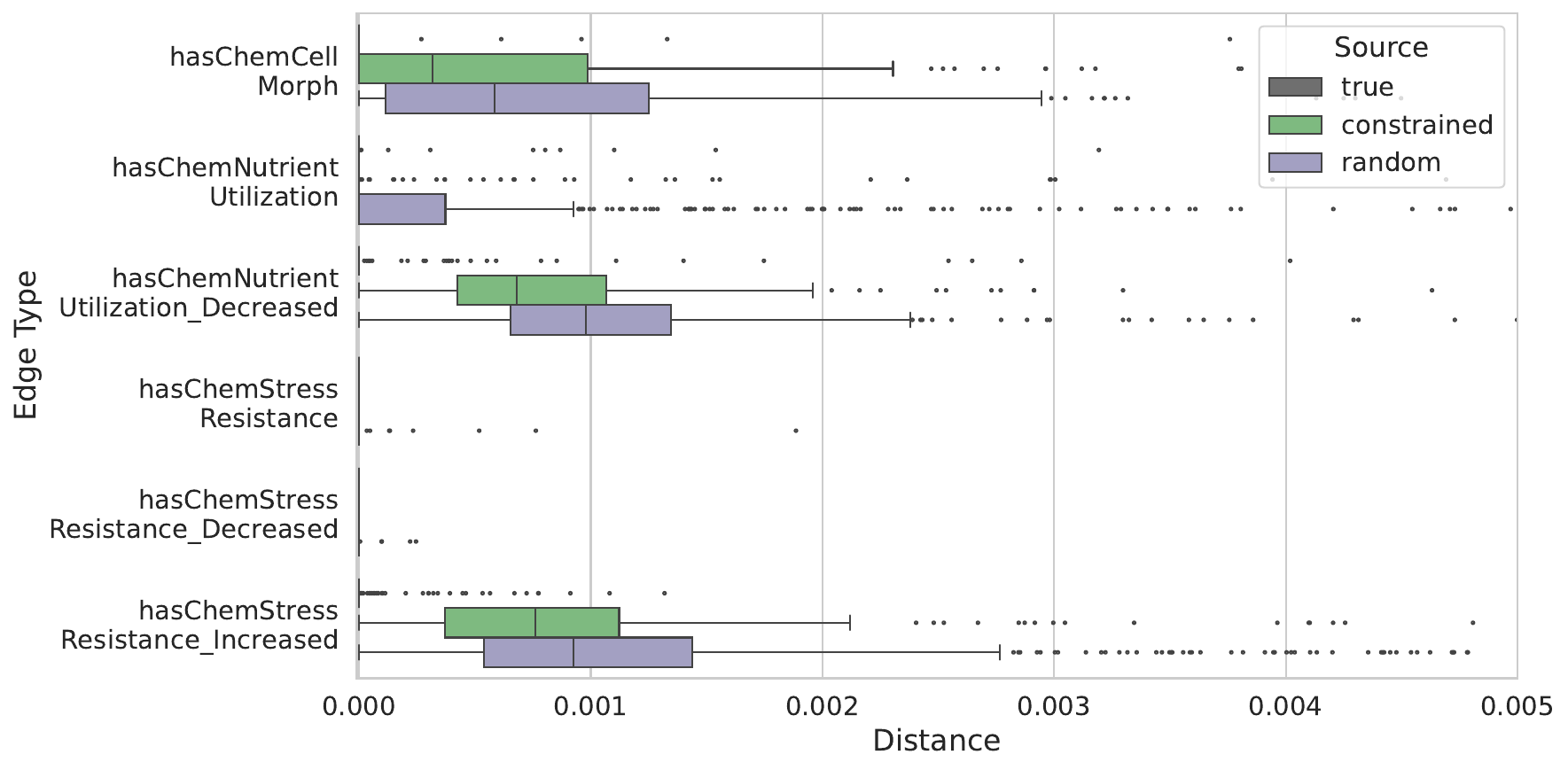}
    \caption{Distribution of distances of box embeddings learned from revised graphs $\tilde{\mathcal{G}}$ to the original embeddings learned from $\mathcal{G}_{\text{train}}$, shown by relation type, for a subset of the relation types in the graph. (The method for calculating these differences is described in \ref{sec:link_eval_method}). For most relations, the distance ranks of randomly drawn edges (constrained to the appropriate class) were significantly different to the ranks of the test edges ($p$ $<$0.05, two-sided Mann--Whitney~U test). Embeddings were learned with inclusion losses.}
    \label{fig:link-evaluation}
\end{figure}

The median distances and loss changes for individual edge revisions to the graph were small, and these measures overall had very large variance. However there were signs in these data which suggest that using these values could be used to rank candidate revisions to a knowledge base. For the majority of the relations, when constraining the random draw to appropriate classes, the distance rank distribution of these randomly drawn edges was significantly different to the rank distribution of the test edges ($p$ $<$0.05, two-sided \nobreak{Mann--Whitney}~U~test). Figure~\ref{fig:link-evaluation} shows a subset of the relation types in the graph. For some relation types, as is the case for `\texttt{hasChemStressResistance}' and \texttt{hasChemStressResistance\_Increased}, there is no appreciable difference in the embeddings after the new edges are added. Summary statistics for these distance evaluations are provided in \tableref{tab:link-eval-per-edge} in Appendix~\ref{apd:link-eval}.


\section{Discussion}
\label{sec:discussion}


In this work we have presented a method generating KG embeddings using GNNs, taking hierarchical class information as well as graph structure into account. We do this by introducing a semantic loss term to the training acting on box transformations of the node embeddings. We have seen that it can be used on its own to generate KG embeddings adhering to subsumptions defined in ontologies to some degree, but more importantly this method shows promise when used together with another, task-specific prediction loss.

The main advantage of our approach is that signals from semantic information encoded in class hierarchies, and signals from predictive tasks can jointly be used to train graph neural networks.

We demonstrated the power of using both these signals together by predicting
digenic deletion fitness from a KG describing \emph{S.~cerevisiae} genes which we constructed. While the predictive $R^2$ of 0.377 may seem low, biological data is inherently noisy, and even replicating experiments is challenging~\citep{roper_testing_2022}. Moreover, our model predicts quantitative outcomes from high-level qualitative information. What is more interesting is how the prediction performance was improved by introducing more hierarchical information to the models. One explanation we envision for the improved performance is that enforcing the class hierarchies has a regularising effect, while also providing semantic grounding for the modelled concepts.
A weakness of the ontologies used, for example ChEBI and GO, is that they do not contain axioms for disjointness between concepts (such as those we introduced in Section~\ref{sec:demo-molecular-function}). This is possibly a consequence of them having been designed primarily for constructing databases, rather than for automated reasoning.
Despite this, the improved performance on the prediction task also suggests that the ontologies used are, at least somewhat, good models of the domains.

KG embeddings that, at least to some extent, adhere to their underlying ontologies can potentially be used for several tasks, even if they were trained with a particular problem in mind. Our trigenic gene deletion experiments are one example of applying the model slightly outside its original domain. The increased performance in this task compared to the digenic deletion is likely, at least partly, due to the individual genes involved no longer being unseen during training. The fitness will depend heavily on the traits of the individual genes, which will be better represented for genes in the training data. The embeddings could potentially be applied to a broader range of tasks, such as GO annotation of genes, which is typically addressed by integrating multiple knowledge sources~\citep{merino_hierarchical_2022}.

We have not utilised any sequence information for our fitness predictions, despite it being the most informative data about genes and fully available for \emph{S.~cerevisiae}. Representing the initial gene embeddings as some encoding of their sequence would provide richer and more meaningful gene embeddings and most likely result in better predictions. However, our current setup will put more emphasis on using the information in the KG as the basis of the predictions. In this way this helps demonstrate both the knowledge in the KG and the usefulness of our embedding method.


The capability of making predictions from qualitative facts enabled interpretability techniques to guide experiment selection, underscoring the value of structured data representation and computational methods in accelerating research. Our edge filtering for viable experiments introduces biases regarding the type of hypotheses generated. Leveraging large language models could be one approach to automatically refine this selection and reveal overlooked experiments.


We suggested two different loss functions for learning box embeddings. The first is based on the volume of overlap between boxes, rewarding overlap for subclasses, and penalising overlaps in the case of disjointness. The second was based on the distance from the boxes fulfilling the subsumption axioms. Studying the 2-dimensional embeddings of in \figureref{fig:trained-boxes-comparison}, the embedding learnt through the $overlap$ loss seems to have slightly better captured the semantics of the ontology. Boxes for subclasses of `\texttt{structural molecule activity}' and `\texttt{molecular function regulator activity}' are mostly contained within their respective superclass, but there were some inconsistencies with the hierarchical structure of the ontology, especially after the message passing of the GNN.
The embedding learnt through the $distance$-based loss instead places the boxes for each class along a diagonal.

Another property of the embeddings learned in this example is the variation in position among instances, which is better for the $overlap$-loss. \figureref{fig:final-boxes-distance} also shows some variation, primarily among instances of `\texttt{small molecule sensor activity}', but there are some large disagreements with the semantics of the ontology. It is probably the case that a 2-dimensional embedding is not sufficient to capture the complexity of this domain.

Interestingly, when class hierarchies were enforced in the gene deletion fitness predictions, the $distance$-based loss yielded better predictive performance. One potential explanation for this finding is that the $distance$-based loss may be particularly well suited as a semantic loss to complement a task-specific loss. By introducing a semantic loss component to our total loss, we of course want to capture how faithfully a given embedding adheres to the semantics of the source ontology. But to have smooth training, a desirable feature of a semantic loss measure is that its gradients are informative when the constraints are not fulfilled.
This is exactly what $\mathcal{L}_{distance}$ does. To obtain useful gradients with $\mathcal{L}_{overlap}$, one could for example use Gumbel boxes as in \cite{dasgupta_improving_2020}, but a result of the introduced smoothing is that losses can remain nonzero even when the semantic constraints are fulfilled. With another loss term primarily guiding the training, in this case $\mathcal{L}_{\text{MSE}}$, the issue of $\mathcal{L}_{distance}$ not discriminating between classes that we observed in Section~\ref{sec:cell-comp-demo} is not as pressing, as the primary loss will likely also push towards being able to discriminate.

Our proposed method for evaluating link revisions to a KG can be seen as an interesting application and direction for future research. Evaluating only the distance in the generated embeddings is, in this setting, not enough to discriminate between true and random edges. It could possibly work better for a more heterogeneous graph, with a richer class hierarchy. A measure of distance, combined with the semantic losses, could represent a measure of surprise. In a scientific discovery context, surprise can be used as part of an active learning algorithm, where a learning agent selects hypotheses that are expected to generate the highest amount of information. This opens up opportunities to create and evaluate scientific hypotheses based on the effect they have on KG embeddings in the domain.

\subsection*{Future work}
We identify several directions for future work based on this paper. First, the hierarchy-aware GNN approach should be evaluated on a broader range of problems, for example tasks from the Open Graph Benchmark. While the results presented here indicate promise for a specific link-level prediction task on a particular KG, we do not assess how well the approach generalises to other settings. We expect that its effectiveness will depend on the availability and informativeness of class hierarchies, as well as on the extent to which they provide information not already encoded in the graph structure.

Representing all graph elements as TBox axioms constitutes a simplification of the underlying semantics. For example, modelling individuals as boxes allows intersections between instances, which lack a clear semantic interpretation. In contrast, instances in knowledge graph embedding models are more commonly represented as points. Similarly, existential restrictions between classes are currently modelled in the same manner as relations between instances, whereas such restrictions could potentially be handled more directly within the semantic loss formulation.

The box representation could also be more tightly integrated into the message-passing process through box-level message aggregation, as in \cite{lin_when_2024}, where messages are propagated via geometric composition of boxes. For hierarchy-aware GNNs on knowledge graphs, this could enable message passing that preserves class hierarchies and ontological constraints directly in the embedding space, rather than enforcing them only through auxiliary loss terms.

We present a method for generating scientific hypotheses and demonstrate its potential through the successful experimental validation of one such hypothesis. Although promising, this constitutes only an initial proof of concept, and substantially more extensive evaluation is required to assess the reliability and broader applicability of the approach. In addition to experimental correctness, future work should evaluate generated hypotheses according to their scientific interest, novelty, and relevance as perceived by domain experts.

Finally, the link evaluation method introduced in Section~\ref{sec:link_eval_method}, while rudimentary in the context of link prediction, may provide an alternative mechanism for hypothesis evaluation. By evaluating sets of added links jointly, hypotheses can be assessed not only in isolation but also in terms of their global impact on the learned representation. Changes in the embedding space can be interpreted as a measure of surprise, while corresponding changes in the semantic loss provide a quantitative signal of consistency with existing knowledge. Together, these signals offer a principled basis for evaluating hypotheses with respect to both novelty and plausibility.




\section{Conclusion}
In this work we have presented a method generating KG embeddings using GNNs, taking hierarchical class information as well as graph structure into account. We show that enforcing the class hierarchies as semantic losses throughout the model can help predictive performance while also producing internal representations which better correspond to our knowledge of the domain. This is demonstrated on a KG we have created from publicly available data about the yeast \emph{S.~cerevisiae}. Based on this KG we can, not only predict biological measurements, but also use interpretability tools to form a hypothesis about phenotype interactions. One such hypothesis was tested and supported by performing a biological experiment, uncovering an association between inositol utilisation and NaCl stress. This illustrates how models with semantic grounding can help in scientific discovery.


The code and data for this project are available at \url{https://github.com/filipkro/kg-box-emb}.

\acks{The computations were enabled by resources provided by the National Academic Infrastructure for Supercomputing in Sweden (NAISS), partially funded by the Swedish Research Council through grant agreement no. 2022-06725. This work was supported by the Wallenberg AI, Autonomous Systems and Software Program (WASP) funded by the Knut and Alice Wallenberg Foundation, the UK Engineering and Physical Sciences Research Council (EPSRC) [EP/R022925/2, EP/W004801/1 and EP/X032418/1], the Chalmers AI Research Centre, and Swedish Research Council Formas [2020-01690].}

\clearpage
\bibliography{main}
\clearpage

\appendix
\section{Examples of description logic in KG}\label{apd:dl_kg}


A description logic example of how a phenotype with a qualifier and chemical are specified in the KG. This example is about decreased (\texttt{APO\_0000003}) utilisation of carbon source (\texttt{APO\_0000096}) of lactate (\texttt{CHEBI\_16004}), observed for the gene `\texttt{YBL030C}'. The gene is linked to the phenotype by the has phenotype relation (\texttt{RO\_0002200}) and to lactate with \texttt{hasChemNutrientUtilization\_Decreased}.

\begin{align}
  \begin{split}
  &\mathtt{APO\_0000098}\text{-}\mathtt{APO\_0000003}\text{-}\mathtt{CHEBI\_16004} \sqsubseteq \mathtt{APO\_0000098} \sqcap \mathtt{APO\_0000003} \\
  &\mathtt{APO\_0000098}\text{-}\mathtt{APO\_0000003}\text{-}\mathtt{CHEBI\_16004} \sqsubseteq \exists \mathtt{aboutChemical}.\mathtt{CHEBI\_16004}\\
  &\mathtt{YBL030C} \sqsubseteq \exists \mathtt{RO\_0002200}.\mathtt{APO\_0000098}\text{-}\mathtt{APO\_0000003}\text{-}\mathtt{CHEBI\_16004} \\
  &\mathtt{YBL030C} \sqsubseteq \exists \mathtt{hasChemNutrientUtilization\_Decreased}.\mathtt{CHEBI}\mathtt{\_16004}.
  \end{split}
  \label{eq:dl_pheno}
\end{align}

A description logic example of how a gene (\texttt{YCR073C}) is positively regulating the protein activity (\texttt{INO\_0000104}) of another gene (\texttt{YLR113W}). This regulation happens during (\texttt{RO\_0002092}) cellular response to heat (\texttt{GO\_0034605}).

\begin{align}
  \begin{split}
  &\mathtt{YCR073C}\text{-}\mathtt{YLR113W}\text{-}\mathtt{protein\_activity}\text{-}\mathtt{positive} \sqsubseteq \mathtt{INO\_0000104} \\
  &\mathtt{YCR073C} \sqsubseteq \exists\mathtt{positive\_regulator\_of}.\mathtt{YCR073C}\text{-}\mathtt{YLR113W}\text{-}\mathtt{protein\_activity}\text{-}\mathtt{positive} \\
  &\mathtt{positive\_regulator\_of}.\mathtt{YCR073C}\text{-}\mathtt{YLR113W}\text{-}\mathtt{protein\_activity}\text{-}\mathtt{positive} \\ 
  & \qquad \qquad \sqsubseteq \exists\mathtt{regulated\_gene}.\mathtt{YLR113W} \\
  &\mathtt{positive\_regulator\_of}.\mathtt{YCR073C}\text{-}\mathtt{YLR113W}\text{-}\mathtt{protein\_activity}\text{-}\mathtt{positive} \\ 
  & \qquad \qquad \sqsubseteq \exists\mathtt{RO\_0002092}.\mathtt{GO\_0034605} \\
  &\mathtt{YCR073C} \sqsubseteq \exists\mathtt{positively\_regulating}.\mathtt{YLR113W}.
  \end{split}
  \label{eq:dl_reg}
\end{align}

\clearpage
\section{Hyperparameters}\label{apd:hyper}

\subsection{Box embedding parameters}\label{apd:box_hyper}
\begin{table}[ht]
  \centering
  \caption{Parameters used for the box embeddings of the different domains}
  \vspace{1em}
  \label{tab:box_hyper}
  \begin{tabular}{|r|r|r|r|r|r|r|}
    \hline
    Domain & Dimensions & Epochs & Lr & Regularisation & \specialcell[t]{Gumbel \\ temperature} & \specialcell[t]{Neg. ex. \\ ratio} \\ \hline
    \specialcell{Material \\ entity} & 10 & 1,000 & 1e-2 & 1e-3 & 0.25 & 2.0 \\ \hline
    \specialcell{Genes} & 8 & 1,000 & 1e-2 & 1e-3 & 0.25 & 4.0 \\ \hline
    \specialcell{Regulations} & 5 & 800 & 1e-2 & 1e-3 & 0.25 & 2.0 \\ \hline
    \specialcell{Molecular \\ functions} & 5 & 800 & 1e-2 & 1e-3 & 0.25 & 2.0 \\ \hline
    \specialcell{Biological \\ processes} & 5 & 800 & 1e-2 & 1e-3 & 0.25 & 2.0 \\ \hline
    \specialcell{Phenotypes} & 5 & 800 & 1e-2 & 1e-3 & 0.25 & 2.0 \\ \hline
    \specialcell{Reactions \& \\ Pathways} & 5 & 800 & 1e-2 & 1e-3 & 0.25 & 2.0 \\ \hline
    \specialcell{Cellular \\ components} & 5 & 800 & 1e-2 & 1e-3 & 0.25 & 2.0 \\ \hline
  \end{tabular}
\end{table}

\FloatBarrier
\subsection{Prediction models}\label{apd:model_hyper}
The best performing model was trained for 160 epochs, with a learning rate of 1e-4, and L2 regularisation weight of 0.1. The depth of the GNN was 2 and the embedding dimensions for the domains are listed in Table \ref{tab:gnn_dims} and are the same throughout the GNN. The fully connected neural network predicting the interaction from the embeddings is of depth 2 with 64, and 1 neurons respectively. For models trained with the semantic loss in \eqref{eq:sem_loss} we used $\alpha = 0.1$ and $\beta = 0.05$.

\begin{table}[ht]
  \centering
  \caption{Embedding dimensions for the different domains throughout the GNN.}
  \label{tab:gnn_dims}
  \vspace{1em}
  \begin{tabular}{|l|c|c|c|}
    \hline
    \specialcell[c]{Embedding \\ dimensions} & 32 & 64 & 128 \\ \hline
    Domains &
    \specialcell[t]{Cellular components \\ Molecular functions \\ Reactions \\ Regulations} &
    \specialcell[t]{Biological processes \\ Phenotypes \\ } &
    \specialcell[t]{Material entities \\ Genes} \\ \hline
  \end{tabular}
\end{table}

\FloatBarrier
\subsection{Box embeddings for training without prediction task}\label{apd:demo_hyper}

\begin{table}[H]
  \centering
  \caption{The embedding models, trained with both $\mathcal{L}_{distance}$ and $\mathcal{L}_{overlap}$ used the same hyperparameters. The learning rate was after each epoch multiplied with $(1 - \text{Lr decay})$ and the regularisation used is presented in \eqref{eq:reg_small_box}, penalising small boxes, with $l_0 = 1$. $\lambda_{s}$, $\lambda$, $\beta$, and $\gamma$ refer to weights in the losses in \eqref{eq:demo_box_loss} and \eqref{eq:demo_box_negative_loss}.}
  \vspace{1em}
  \label{tab:family_box_hyper}
  \begin{tabular}{|r|r|r|r|r|r|r|}

    \hline
    Epochs & Initial lr & Lr decay & \specialcell[t]{Regular- \\ isation, $\lambda$} & \specialcell[t]{Small Box \\ Regular- \\ isation, $\lambda_{s}$} & \specialcell[t]{Negative \\ weight, $\beta$} & \specialcell[t]{Negative \\ weight, $\gamma$} \\ \hline
    500 & 0.1 & 0.001 & 0.001 & 0.01 & 0.5 & 1.0 \\ \hline

  \end{tabular}
\end{table}

\clearpage
\section{Prediction model comparisons}
\label{apd:pred_performance}

\begin{table}[ht]
  \centering
  \caption{The architecture described in Section~\ref{sec:gnns} with different levels of hierarchy integration. These results are also presented in \tableref{tab:results}. Results from 10-fold cross-validation of digenic deletion fitness.}
  \label{tab:gnn_performance}
  \begin{tabular}{|l|l|l|l|} \hline
    & Description & Mean $R^2$ & SD \\ \hline
    $c$ & GNN without box embeddings & 0.348 & 0.050 \\ \hline
    $d$ & GNN with \texttt{subClassOf}-links in KG & 0.350 & 0.049 \\ \hline
    $e$ & GNN with prior box embeddings & 0.360 & 0.043 \\ \hline
    $f$ & GNN with prior box embeddings + $\mathcal{L}_{overlap}$ & 0.368 & 0.038  \\ \hline
    $g$ & GNN with prior box embeddings + $\mathcal{L}_{distance}$ & \textbf{0.377} & 0.046 \\ \hline
  \end{tabular}
\end{table}

\begin{table}[ht]
  \centering
  \caption{Baseline performance. Results from 10-fold cross-validation of digenic deletion fitness.}
  \label{tab:baselines}
  \begin{tabular}{|l|l|l|l|} \hline
    & Description & Mean $R^2$ & SD \\ \hline
    $a$ & ComplEx + LightGBM & 0.191 & 0.039 \\ \hline
    $b$ & Instantiations + LightGBM & 0.211 & 0.022 \\ \hline
    $h$ & ComplEx + MLP & 0.165 & 0.032 \\ \hline
    $i$ & Box$^2$EL + LightGBM & 0.016 & 0.007 \\ \hline
  \end{tabular}
\end{table}

\begin{table}[ht]
  \centering
  \caption{Impact on predictive performance when ignoring rare edges in the graph. Results from 10-fold cross-validation of digenic deletion fitness.}
  \label{tab:ignored_edges}
  \begin{tabular}{|l|l|l|l|} \hline
    & Description & Mean $R^2$ & SD \\ \hline
    $j$ & Ignoring edges with fewer than 100 examples & 0.333 & 0.043 \\ \hline 
    $k$ & Ignoring edges with fewer than 500 examples & 0.365 & 0.045 \\ \hline
    $g$ & Ignoring edges with fewer than 1,000 examples & \textbf{0.377} & 0.046 \\ \hline
    $l$ & Ignoring edges with fewer than 5,000 examples & 0.368 & 0.045 \\ \hline
  \end{tabular}
\end{table}

\begin{table}[ht]
  \centering
  \caption{Impact on predictive performance when combining gene embeddings using the methods presented in \tableref{tab:gene_combination}. Results from 10-fold cross-validation of digenic deletion fitness.}
  \label{tab:gene_combine_performance}
  \begin{tabular}{|l|l|l|l|} \hline
    & Description & Mean $R^2$ & SD \\ \hline
    $g$ & Element-wise product & \textbf{0.377} & 0.046 \\ \hline
    $m$ & Bilinear & 0.363 & 0.042 \\ \hline 
    $n$ & Concatenation & 0.349 & 0.062 \\ \hline
    $o$ & Intersection & 0.347 & 0.039 \\ \hline 
  \end{tabular}
\end{table}

\begin{table}[ht]
  \centering
  \caption{Impact on predictive performance when varying the dimensionalities of the embeddings of the model. Varying initial dimensions means the box embeddings in \tableref{tab:box_hyper} are used. For the same initial dimensions, box embeddings in 10 dimensions are used, found with the hyperparameters in \tableref{tab:box_hyper}. Varying GNN dimensions refers to the parameters in \tableref{tab:gnn_dims}. For the same GNN dimensions all embeddings are in 128 dimensions. Results from 10-fold cross-validation of digenic deletion fitness.}
  \label{tab:gnn_dimensions}
  \begin{tabular}{|l|l|l|l|} \hline
    & Description & Mean $R^2$ & SD \\ \hline
    $g$ & Varying initial dimensions, varying GNN dimensions & \textbf{0.377} & 0.046 \\ \hline
    $p$ & Varying initial dimensions, same GNN dimensions (128) & 0.361 & 0.046 \\ \hline 
    $q$ & Same initial dimension (10), same GNN dimensions (128) & 0.372 & 0.053 \\ \hline
    $r$ & Same initial dimensions (10), varying GNN dimensions & 0.368 & 0.046 \\ \hline
  \end{tabular}
\end{table}

\begin{table}[ht]
  \centering
  \caption{Comparison of different model architectures. Model $f$ refers to the architecture presented in Section~\ref{sec:gnns}, $r$ uses a Graph Attention Network (GATv2) \citep{brody2022how}, and $s$ uses the same architecture as $f$, but do not split the graph into distinct domains which are embedded separately. Results from 10-fold cross-validation of digenic deletion fitness.}
  \label{tab:gnn_architecture}
  \begin{tabular}{|l|l|l|l|} \hline
    & Description & Mean $R^2$ & SD \\ \hline
    $g$ & GraphSAGE, NN prediction head, split domains & \textbf{0.377} & 0.046 \\ \hline
    $s$ & GATv2, NN prediction head, split domains & 0.253 & 0.030 \\ \hline
    $t$ & GraphSAGE, NN prediction head, one domain & 0.128 & 0.184 \\ \hline 
  \end{tabular}
\end{table}

\clearpage
\section{Model-driven experiment}
\label{apd:experiment}

\subsection{Edge filtering}
\label{apd:exp_filtering}
\begin{table}[ht]
\centering
\caption{We filter for co-occurring edge pairs in which at least one edge connects a gene to a node that is a subclass of one of the following APO classes, related to nutrient utilisation.}
\begin{tabular}{|ll|}
\hline
\textbf{APO Class} & \textbf{Description} \\ \hline
\texttt{APO\_0000096} & General nutrient utilisation \\ \hline
\texttt{APO\_0000097} & Auxotrophy \\ \hline
\texttt{APO\_0000099} & Utilisation of nitrogen source \\ \hline
\texttt{APO\_0000100} & Nutrient uptake \\ \hline
\texttt{APO\_0000125} & Utilisation of phosphorous source \\ \hline
\texttt{APO\_0000219} & Utilisation of sulfur source \\ \hline
\end{tabular}
\label{tab:apo_nutrient_classes}
\end{table}

\begin{table}[ht]
\centering
\caption{We also allow edge pairs where at least one of the edges links a gene to a chemical through any of the following relations.}
\begin{tabular}{|l|}\hline
\texttt{hasChemNutrientUtilization}\\
\texttt{hasChemNutrientUtilization\_Increased}\\
\texttt{hasChemNutrientUtilization\_Decreased} \\ \hline
\end{tabular}
\label{tab:apd_allowed_edge}
\end{table}
\newpage
\FloatBarrier
\subsection{Top edge pairs}
\label{apd:top_edges}
\begin{table}[ht]
  \centering
  \caption{The 10 edge pairs with the highest importance weight after filtering for the criteria specified in Appendix~\ref{apd:exp_filtering}. The edge pair selected for the experiment is highlighted. \texttt{Ch.Nutr.Util.} is short for `\texttt{hasChemNutrientUtilization}', \texttt{Ch.Nutr.Util.Dec.} is short for \texttt{hasChemNutrientUtilization\_Decreased}, and \texttt{Ch.StressRes.} is short for `\texttt{hasChemStressResistance}'. Clarifications of terms can be found in \tableref{tab:speaking_terms}.}
  \label{tab:top_edges}
  \begin{tabular}{|c|l|l|l|l|}
    \hline
    Importance & Relation1 & Class1 & Relation2 & Class2 \\ \hline
    0.003471 & \texttt{Ch.Nutr.Util.} & \texttt{CHEBI\_17268} & \texttt{Ch.StressRes.} & \texttt{CHEBI\_22907} \\ \hline
    0.002002 & \texttt{Ch.StressRes.} & \texttt{CHEBI\_22907} & \texttt{has\_phenotype} & \specialcell{\texttt{APO\_0000099-}\\\texttt{APO\_0000245-}\\\texttt{CHEBI\_14321}} \\ \hline 
    \rowcolor{SeaGreen3!30!} \textbf{0.001985} & \texttt{\textbf{Ch.Nutr.Util.}} & \textbf{\texttt{CHEBI\_17268}} & \textbf{\texttt{Ch.StressRes.}} & \textbf{\texttt{CHEBI\_26710}} \\ \hline
    0.001705 & \texttt{Ch.Nutr.Util.Dec.} & \texttt{CHEBI\_23414} & \texttt{Ch.StressRes.} & \texttt{CHEBI\_22907} \\ \hline
    0.001679 & \texttt{hasChemCellMorph} & \texttt{CHEBI\_26710} & \texttt{has\_phenotype} & \specialcell{\texttt{APO\_0000099-}\\\texttt{APO\_0000245-}\\\texttt{CHEBI\_14321}} \\ \hline
    0.001580 & \texttt{Ch.Nutr.Util.} & \texttt{CHEBI\_17268} & \texttt{Ch.StressRes.} & \texttt{CHEBI\_50145} \\ \hline
    0.001541 & \texttt{Ch.Nutr.Util.Dec.} & \texttt{CHEBI\_77995} & \texttt{Ch.StressRes.} & \texttt{CHEBI\_49470} \\ \hline
    0.001537 & \texttt{Ch.StressRes.} & \texttt{CHEBI\_22907} & \texttt{has\_phenotype} & \specialcell{\texttt{APO\_0000099-}\\\texttt{APO\_0000245-}\\\texttt{CHEBI\_26271}} \\ \hline
    0.001500 & \texttt{Ch.Nutr.Util.} & \texttt{CHEBI\_17268} & \texttt{has\_phenotype} & \specialcell{\texttt{APO\_0000059-}\\\texttt{APO\_0000002-}\\\texttt{CHEBI\_26710}} \\ \hline
    0.001477 & \texttt{Ch.Nutr.Util.Dec.} & \texttt{CHEBI\_16236} & \texttt{Ch.StressRes.} & \texttt{CHEBI\_22907} \\ \hline
\end{tabular}
\end{table}

\begin{table}[ht]
  \centering

  \caption{Clarifications for terms in the \tableref{tab:top_edges}.}
  \vspace{1em}
  \label{tab:speaking_terms}
  \begin{tabular}{|l|l|}\hline
    Identifier & Label \\ \hline
    \texttt{CHEBI\_17268} & Myo-inositol \\ \hline
    \texttt{CHEBI\_22907} & Bleomycin \\ \hline
    \texttt{APO\_0000099} & Util. of nitrogen source \\ \hline
    \texttt{APO\_0000245} & Decreased \\ \hline
    \texttt{CHEBI\_14321} & Glutamate \\ \hline
    \texttt{CHEBI\_26710} & Sodium chloride \\ \hline
    \texttt{CHEBI\_23414} & Cpper sulfate \\ \hline
    \texttt{CHEBI\_50145} & Fenpropimorph \\ \hline
    \texttt{CHEBI\_77995} & Diphenyl, phenanthroline \\ \hline
    \texttt{CHEBI\_26271} & Proline \\ \hline
    \texttt{APO\_0000059} & Vacuolar morphology \\ \hline
    \texttt{APO\_0000002} & Abnormal \\ \hline
    \texttt{CHEBI\_16236 } & Ethanol \\ \hline
\end{tabular}
  
\end{table}

\FloatBarrier
\subsection{Cultivation method}
\label{apd:cultivation}
The $\Delta$\emph{ino1} deletion mutant was taken from the EUROSCARF deletion collection, with the strain background being BY4741, genotype: MATa, \emph{his3}$\Delta$1, \emph{leu2}$\Delta$0, \emph{met15}$\Delta$0, \emph{ura3}$\Delta$0 (Y01272).
 
The $\Delta$\emph{ino1} mutant was pre-cultured overnight in minimally buffered delft media containing the following: 5g/L (NH4)2SO4, 3g/L KH2PO4, 0.5g/L MGSO4. 7H2O, and 1mL/L trace metal and vitamin solutions as described by \cite{verduyn_effect_1992}, 25 mg/L myo-inositol and 2\% glucose (w/v) in 30$^\circ$C, and 220rpm.
The pre-culture was adjusted to 0.5 OD600, and robotically dispensed with a 1:20 dilution into a 96-well microculture plate using a Hamilton Microlab Star liquid handling robot. A negative control  was also included to assess the baseline growth of the $\Delta$\emph{ino1} mutant without any supplementation of myo-inositol. Additionally, myo-inositol-free media with 0.25\% (w/v) glucose, myo-inositol (Sigma aldrich 57570-100G), Sodium chloride (Merck 1064041000) and MilliQ-water was robotically dispensed, resulting in a total volume of 250$\mu$L and the concentrations defined in Table~\ref{tab:concentrations}.

\begin{table}[H]
  \centering
  \caption{The concentrations of inositol and NaCl used for the experiment.}
  \label{tab:concentrations}
  \vspace{0.5em}
  \begin{tabular}{|l|l|}
    \hline
    Inositol & NaCl \\ \hline
    0.00 $m$Molar & 0.0 Molar \\ \hline
    0.01 $m$Molar & 0.0 Molar \\ \hline
    0.01 $m$Molar & 0.3 Molar \\ \hline
    0.01 $m$Molar & 0.6 Molar \\ \hline
    0.05 $m$Molar & 0.0 Molar \\ \hline
    0.05 $m$Molar & 0.3 Molar \\ \hline
    0.05 $m$Molar & 0.6 Molar \\ \hline
    0.25 $m$Molar & 0.0 Molar \\ \hline
    0.25 $m$Molar & 0.3 Molar \\ \hline
    0.25 $m$Molar & 0.6 Molar \\ \hline
    
  \end{tabular}
  
\end{table}
 
A robust plate layout was generated with PLAID~\citep{francisco_rodriguez_designing_2023}. The processed plate was cultivated in the automated laboratory cell Eve. The plate was transferred from an automated incubator (30$^\circ$C) to a Teleshaker Magnetic Shaking System, where it was shaken for 30s at 800 rpm, divided evenly between clockwise and counter-clockwise double-orbital shaking. After shaking, the plate was transferred to a BMG Polarstar plate reader, where it underwent optical density measurements at 600 nM (the temperature in the plate reader was kept at a constant 30$^\circ$C). After measuring, the plate was returned to the incubator. The protocol was automatically repeated every 20 min for up to 24 h.

\FloatBarrier
\subsection{Growth data processing and statistical testing}
\label{apd:stats}
\begin{figure}[H]
  \centering
  \includegraphics[width=0.8\textwidth]{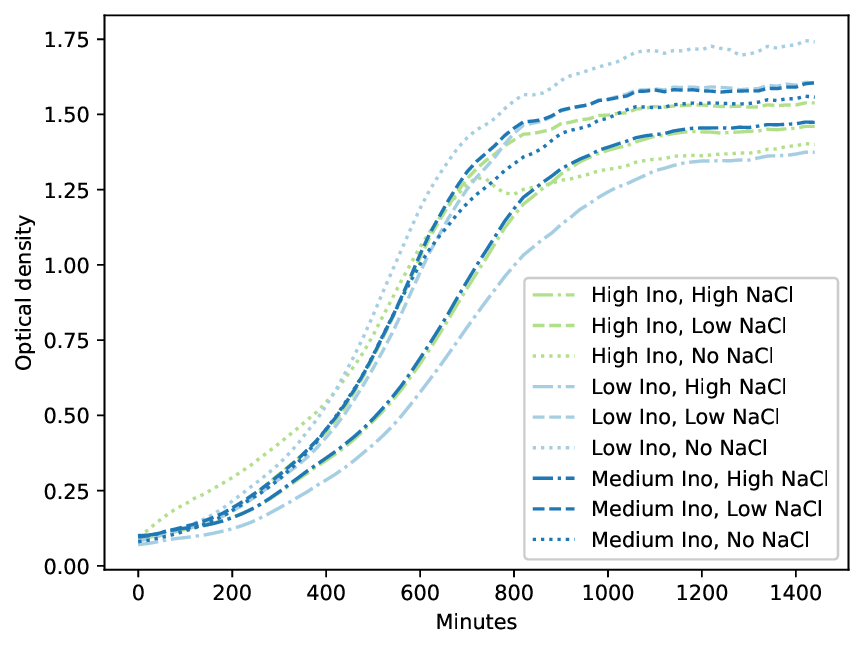}
  \caption{Growth curves showing the mean optical densities of the 6-8 repetitions for the different experimental groups. Optical density (at 600nM) is a unitless measurement typically used as an indirect measure of cell density and biomass.}
  \label{fig:growth_curves}
\end{figure}

Outliers in the growth curves (measured through optical density at 600nm) were identified and filtered using the interquartile range (IQR), where any data points outside the range of [Q1-1.5 IQR, Q3+1.5 IQR] were excluded from the dataset. The filtered curves were then subsequently smoothed using a rolling mean of window size 3. The resulting averaged growth curves can be seen in \figureref{fig:growth_curves}. Area under curve was calculated using \texttt{numpy.trapz} (\texttt{v1.26.4}). To assess the effects of inositol and NaCl on AUC, a generalised linear model was employed (\texttt{statsmodels v0.14.4}). The model was fitted using a Gaussian family distribution. Choice $\alpha$-value was set at 0.05. We modelled all factors as categorical to avoid imposing any assumptions on linearity. The model is specified as follows:

\begin{equation}
  \text{AUC} \backsim C(Inositol) \times C(NaCl).
  \label{apd:eq:stat_model}
\end{equation}

\begin{table}[h]
  \centering
  \caption{Estimated parameters from the GLM examining the effects of myo-Inositol-supplementation and NaCl treatment on growth dynamics. The table presents coefficient estimates, $p$-values and confidence intervals for the main effects and interaction terms. Significant interactions indicate that the effect of myo-inositol supplementation changes depending on treatment levels. The two highlighted rows indicate the significant interaction effect.}
  \label{tab:stats_results}
  \vspace{-0.7em}
  \begin{tabular}{|l|c|c|c|}
    \hline
    & \specialcell{Coefficient\\($\times10^3$)} & \specialcell{Confidence \\interval ($\times10^3$)} & $p$-value \\ \hline
    Intercept & 97.29 & $[$88.20, 106$]$ & 0.000 \\ \hline
    Medium inositol & -11.84 &$[$-24.7, 1.02$]$ & 0.071 \\ \hline
    High inositol & -14.56 & $[$-27.9, -1.24$]$ & \textbf{0.032} \\ \hline
    Low NaCl & -9.50 & $[$-22.4, 3.37$]$ & 0.148 \\ \hline
    High NaCl & -30.61 & $[$-43.5, -17.8$]$ & \textbf{0.000} \\ \hline
    Medium inositol $\times$ Low NaCl & 13.11 & $[$-5.40, 31.6$]$ & 0.165 \\ \hline
    High inositol $\times$ Low NaCl & 13.38 & $[$-5.45, 32.2$]$ & 0.164 \\ \hline
    \rowcolor{SeaGreen3!30!} Medium inositol $\times$ High NaCl & 20.92 & $[$2.41, 39.4$]$ & \textbf{0.027} \\ \hline
    \rowcolor{SeaGreen3!30!} High inositol $\times$ High NaCl & 22.64 & $[$3.40, 41.9$]$ & \textbf{0.021} \\ \hline
  \end{tabular}
\end{table}

\clearpage
\section{Link evaluation results per edge type}\label{apd:link-eval}

\begin{table}[ht]
  \centering
  \tiny
  \caption{Link evaluation results per edge type.}
  \label{tab:link-eval-per-edge}
  \vspace{0.5em}
  \rotatebox{90}{%

  \begin{tabular}{p{3.9cm}rrrrrrrr}
  \toprule
  Edge Type & \specialcell[t]{\#Test\\Edges} & \specialcell[t]{Mean\\Dist.\\(real)} & \specialcell[t]{Mean Dist. \\ (constrained)} & \specialcell[t]{Mean\\Dist.\\(random)} & \specialcell[t]{M-W U\\(Real vs.\\Random)} & \specialcell[t]{p-value\\(Real vs.\\Random)} & \specialcell[t]{M-W U\\(Real vs.\\Constrained)} & \specialcell[t]{p-value\\(Real vs.\\Constrained)} \\
  \midrule
  BFO\_0000050 & 500 & 2.578e-04 & 6.271e-04 & 5.339e-04 & 48417.5 & 2.790e-64 & 82420.0 & 8.367e-22 \\
  BFO\_0000051 & 500 & 2.114e-04 & 1.941e-04 & 2.123e-04 & 129566.0 & 2.239e-01 & 125009.5 & 9.981e-01 \\
  BFO\_0000066 & 154 & 3.423e-04 & 1.812e-03 & 3.121e-03 & 1318.0 & 2.676e-44 & 3515.5 & 3.007e-29 \\
  RO\_0000057 & 268 & 9.756e-04 & 8.101e-04 & 1.568e-03 & 32214.0 & 3.917e-02 & 40732.5 & 7.176e-03 \\
  RO\_0000087 & 500 & 1.487e-07 & 7.363e-08 & 5.085e-08 & 125250.0 & 5.641e-01 & 125250.0 & 5.641e-01 \\
  RO\_0001025 & 500 & 1.882e-06 & 4.538e-05 & 2.369e-04 & 39615.0 & 9.627e-106 & 103855.0 & 5.874e-19 \\
  RO\_0002092 & 500 & 3.008e-06 & 2.505e-03 & 1.581e-03 & 21128.5 & 8.319e-142 & 42882.0 & 5.331e-102 \\
  RO\_0002200 & 500 & 1.465e-05 & 3.496e-04 & 7.656e-04 & 28532.5 & 2.766e-122 & 61920.0 & 3.883e-66 \\
  RO\_0002211 & 500 & 6.292e-05 & 1.723e-04 & 9.614e-04 & 16370.5 & 8.205e-141 & 105081.5 & 6.095e-11 \\
  RO\_0002212 & 500 & 6.503e-04 & 1.363e-03 & 5.717e-04 & 149994.0 & 2.525e-11 & 113514.0 & 6.415e-03 \\
  RO\_0002213 & 500 & 8.264e-04 & 1.319e-03 & 6.458e-04 & 152742.5 & 3.195e-10 & 118608.0 & 1.861e-01 \\
  RO\_0002233 & 500 & 1.520e-03 & 3.474e-03 & 5.530e-03 & 34860.0 & 3.657e-90 & 60270.5 & 1.124e-48 \\
  RO\_0002234 & 500 & 1.755e-03 & 3.260e-03 & 4.422e-03 & 41317.0 & 9.300e-78 & 65275.5 & 1.096e-41 \\
  RO\_0002326 & 123 & 1.689e-04 & 2.285e-04 & 4.176e-04 & 7129.0 & 2.712e-01 & 7034.5 & 1.902e-01 \\
  RO\_0002327 & 500 & 6.910e-05 & 2.628e-04 & 8.388e-04 & 93425.0 & 1.358e-20 & 97765.0 & 3.737e-16 \\
  RO\_0002331 & 500 & 4.410e-08 & 2.594e-06 & 3.656e-04 & 110971.5 & 3.377e-14 & 124999.5 & 1.000e+00 \\
  has\_functional\_parent & 500 & 7.659e-05 & 1.070e-04 & 1.603e-04 & 107705.0 & 2.071e-10 & 119350.5 & 1.271e-02 \\
  has\_parent\_hydride & 365 & 1.983e-04 & 6.620e-04 & 8.974e-04 & 26468.0 & 6.410e-49 & 42928.0 & 1.521e-19 \\
  is\_conjugate\_acid\_of & 500 & 6.061e-05 & 5.847e-05 & 9.118e-05 & 125726.0 & 6.824e-01 & 124774.0 & 9.897e-01 \\
  is\_conjugate\_base\_of & 500 & 9.977e-04 & 9.682e-04 & 1.105e-03 & 124591.0 & 9.287e-01 & 155845.0 & 1.416e-11 \\
  is\_enantiomer\_of & 500 & 1.896e-03 & 1.450e-03 & 1.868e-03 & 136175.5 & 1.440e-02 & 158041.5 & 3.806e-14 \\
  is\_substituent\_group\_from & 258 & 7.831e-04 & 6.807e-04 & 1.012e-03 & 24170.0 & 7.363e-08 & 32937.5 & 8.388e-01 \\
  is\_tautomer\_of & 389 & 2.202e-04 & 2.026e-04 & 2.099e-04 & 74080.5 & 5.379e-01 & 79678.5 & 8.296e-02 \\
  aboutChemical & 500 & 1.244e-05 & 4.705e-05 & 1.061e-03 & 89850.0 & 5.328e-34 & 123477.5 & 2.087e-01 \\
  catalyzedBy & 500 & 5.007e-04 & 8.114e-04 & 1.368e-03 & 70205.0 & 3.460e-33 & 82218.0 & 7.053e-21 \\
  catalyzedByGene & 410 & 3.229e-04 & 5.789e-04 & 8.031e-04 & 54720.5 & 5.104e-18 & 51335.0 & 4.838e-22 \\
  encodedBy & 500 & 9.142e-04 & 3.046e-04 & 8.949e-04 & 192462.0 & 2.198e-49 & 214997.0 & 1.863e-86 \\
  hasChemCellMorph & 500 & 1.386e-05 & 5.905e-04 & 9.649e-04 & 22913.0 & 3.254e-137 & 50092.5 & 6.066e-89 \\
  hasChemNutrientUtilization & 500 & 1.740e-05 & 8.846e-05 & 6.804e-04 & 89428.0 & 4.286e-34 & 119228.0 & 3.233e-04 \\
  hasChemNutrientUtilization\_Decreased & 293 & 1.588e-04 & 1.105e-03 & 1.455e-03 & 3409.0 & 4.242e-91 & 6908.0 & 6.862e-78 \\
  hasChemStressResistance & 500 & 0.000e+00 & 0.000e+00 & 7.486e-06 & 123000.0 & 4.545e-03 & 125000.0 & 1.000e+00 \\
  hasChemStressResistance\_Decreased & 500 & 0.000e+00 & 0.000e+00 & 1.330e-06 & 123750.0 & 2.511e-02 & 125000.0 & 1.000e+00 \\
  hasChemStressResistance\_Increased & 500 & 2.258e-05 & 9.645e-04 & 1.439e-03 & 6243.0 & 3.008e-167 & 21798.5 & 1.434e-133 \\
  hasGeneProductPart & 500 & 8.593e-04 & 9.468e-04 & 1.363e-03 & 108666.5 & 3.481e-04 & 117656.0 & 1.078e-01 \\
  hasRightParticipant & 103 & 2.895e-04 & 1.905e-03 & 2.359e-03 & 1814.0 & 1.259e-16 & 3750.0 & 1.650e-04 \\
  negative\_regulator\_of & 149 & 4.904e-04 & 1.686e-04 & 1.135e-03 & 4601.0 & 2.371e-18 & 19438.0 & 3.664e-29 \\
  negatively\_regulating\_transcription & 218 & 3.719e-04 & 6.226e-04 & 5.868e-04 & 19703.0 & 2.035e-03 & 14560.0 & 2.661e-12 \\
  positive\_regulator\_of & 252 & 5.027e-04 & 2.059e-04 & 1.089e-03 & 20142.0 & 1.233e-12 & 53510.0 & 2.045e-40 \\
  positively\_regulating\_transcription & 376 & 2.299e-04 & 4.084e-04 & 5.691e-04 & 51171.0 & 5.630e-11 & 42113.0 & 8.380e-22 \\
  regulating\_gene & 500 & 9.037e-05 & 1.770e-04 & 8.146e-04 & 19748.5 & 1.030e-118 & 94922.0 & 1.715e-11 \\
  regulating\_protein\_activity & 108 & 5.941e-04 & 6.883e-04 & 1.004e-03 & 6566.0 & 1.102e-01 & 5433.0 & 3.856e-01 \\
  regulating\_transcription & 500 & 3.737e-05 & 1.064e-04 & 8.277e-04 & 6146.0 & 9.529e-153 & 105635.0 & 5.863e-06 \\
  regulator\_of & 500 & 4.440e-04 & 1.253e-04 & 8.625e-04 & 122335.0 & 5.596e-01 & 223930.0 & 4.458e-104 \\
  \bottomrule
  \end{tabular}
  }
\end{table}

\end{document}